\documentclass[prodmode, acmtkdd]{acmsmall} 

\usepackage[ruled]{algorithm2e}
\graphicspath{ {fig/} }
\usepackage{graphicx}
\usepackage{amsmath}
\usepackage{amssymb}
\usepackage{fixltx2e}
\usepackage{algorithmic}

\usepackage{array}
\newcolumntype{L}[1]{>{\raggedright\let\newline\\\arraybackslash\hspace{0pt}}m{#1}}
\newcolumntype{C}[1]{>{\centering\let\newline\\\arraybackslash\hspace{0pt}}m{#1}}
\newcolumntype{R}[1]{>{\raggedleft\let\newline\\\arraybackslash\hspace{0pt}}m{#1}}

\usepackage{tikz}
\def\checkmark{\tikz\fill[scale=0.4](0,.35) -- (.25,0) -- (1,.7) -- (.25,.15) -- cycle;}

\usepackage{booktabs}
\usepackage[flushleft]{threeparttable}

\SetAlFnt{\small}
\SetAlCapFnt{\small}
\SetAlCapNameFnt{\small}
\SetAlCapHSkip{0pt}
\IncMargin{-\parindent}

\acmVolume{9}
\acmNumber{4}
\acmArticle{39}
\acmYear{2017}
\acmMonth{8}

\doi{0000001.0000001}

\issn{1234-56789}

\begin{document}

\markboth{H. R. Bonab and F. Can}{GOOWE: Geometrically Optimum and Online-Weighted Ensemble Classifier}

\title{GOOWE: Geometrically Optimum and Online-Weighted Ensemble Classifier for Evolving Data Streams}
\author{Hamed~R.~Bonab
\affil{Bilkent University}
Fazli~Can
\affil{Bilkent University}}

\begin{abstract}
Designing adaptive classifiers for an evolving data stream is a challenging task due to the data size and its dynamically changing nature. Combining individual classifiers in an online setting, the ensemble approach, is a well-known solution. It is possible that a subset of classifiers in the ensemble outperforms others in a time-varying fashion. However, optimum weight assignment for component classifiers is a problem which is not yet fully addressed in online evolving environments. We propose a novel data stream ensemble classifier, called Geometrically Optimum and Online-Weighted Ensemble (GOOWE), which assigns optimum weights to the component classifiers using a sliding window containing the most recent data instances. We map vote scores of individual classifiers and true class labels into a spatial environment. Based on the Euclidean distance between vote scores and ideal-points, and using the linear least squares (LSQ) solution, we present a novel, dynamic, and online weighting approach. While LSQ is used for batch mode ensemble classifiers, it is the first time that we adapt and use it for online environments by providing a spatial modeling of online ensembles. In order to show the robustness of the proposed algorithm, we use real-world datasets and synthetic data generators using the MOA libraries. First, we analyze the impact of our weighting system on prediction accuracy through two scenarios. Second, we compare GOOWE with 8 state-of-the-art ensemble classifiers in a comprehensive experimental environment. Our experiments show that GOOWE provides improved reactions to different types of concept drift compared to our baselines. The statistical tests indicate a significant improvement in accuracy, with conservative time and memory requirements. 
\end{abstract}

%
%


%
%



\acmformat{Hamed R. Bonab and Fazli Can, 2017. GOOWE: Geometrically Optimum and Online-Weighted Ensemble Classifier for Evolving Data Streams.}

\begin{bottomstuff}
The paper is accepted for publication in the ACM Transactions on Knowledge Discovery from Data (TKDD) in August 2017.\\
The authors are with the Bilkent Information Retrieval Group, Computer Engineering Department, Bilkent University, 06800, Ankara, Turkey.\\
The current address of Hamed R. Bonab: College of Information and Computer Sciences, University of Massachusetts, Amherst, MA 01003, USA.
\end{bottomstuff}

\maketitle

\section{Introduction}
The automation of several processes in daily life has dramatically increased the number of data stream generators. Mining the data generated in real-world applications; like traffic management data, click streams in web exploration, detailed call logs, stock market and business transactions, social and computer network logs, and many other such examples; introduced several challenges to the domain. These challenges are mostly due to the size and time-evolving nature of these data streams. The cost and effort of storing and retrieving this type of data made the on-the-fly real-time analysis of incoming data crucial \cite{bib:gama2010book}. 

In such dynamically evolving and non-stationary environments, data distribution can change over time, this is referred to as concept drift \cite{bib:survey}. However, some of these changes are not real concept drifts, and they do not need to be reacted to by adaptive classifiers. Real concept drift is referred to as change in the conditional distribution of the output, given the input features, while the distribution of the input may stay unchanged \cite{bib:gama2010book,bib:survey}. An example of evolving environments is filtering spam emails, in which the definition of the spam class label may change with time. Since users specify these class labels, and their preferences may also change with time, the conditional distribution of labels for incoming emails can change \cite{bib:kuncheva1}.

Designing a classifier for time-evolving data streams has some considerations to be addressed, compared to traditional classifiers. Since data arrives continuously, any proposed algorithm needs to process it under strict time constraints. Handling large volumes of data in main memory is impractical, so the proposed algorithm must use limited memory. Patterns of change in target concepts are categorized into sudden/abrupt, incremental, gradual, and reoccurring drifts \cite{bib:2009,bib:survey,bib:kuncheva2,bib:surveybifet2017,bib:surveygama2017}. Effective classifiers should be able to handle these concept drifts. 

More recently, many drift-aware adaptive learning algorithms have been developed. Among these algorithms, ensemble methods are naturally more consistent with the needs of the problem, and they are proven to outperform single algorithms statistically and computationally \cite{bib:2009,bib:aue,bib:addExpert,bib:kuncheva1,bib:awe,bib:surveybifet2017,bib:surveygama2017}. It is possible that a subset of classifiers in the ensemble outperforms others in a time-varying fashion. However, optimum weight assignment for component classifiers is a problem which is not yet fully addressed in online evolving environments \cite{bib:sampling}. We propose a novel data stream ensemble classifier which assigns optimum weights to the component classifiers using a sliding window containing the most recent data instances. Since ensemble methods use individual classifiers inside their models, this does not decrease the importance of designing more adaptive individual classifiers for evolving data streams. Improving the performance of individual classifiers in terms of accuracy and resource usage can also increase the performance of an ensemble.

In this article, we concentrate on designing a geometric framework for dynamic weighting of component classifiers for ensemble methods. We model our ensemble in a spatial environment and use the Euclidean distance as our measure of closeness. We try to find an optimum weighting function based on LSQ, leading to a system of linear equations which describes the ensemble more precisely. Based on this system of linear equations, we design our algorithm called Geometrically Optimum and Online-Weighted Ensemble (GOOWE)---pronounced gooey (/'g\"{u}-\={e}/). It is inspired from the geometry of voting, which is a well-known domain in the political and social sciences, and economics. The geometric analysis of individual votes for aggregation is proven to outperform existing solutions in these fields. In aggregation, various rules may have conflicting votes, i.e., ``the paradox of voting.'' Finding classes of profiles, uncovering paradoxes, and determining the likelihood of disagreements are among the problems addressed by the geometry of voting \cite{bib:saari2008}.

\begin{table}
	\tbl{Symbol Notation\label{tab:notation}}
	{
		\begin{tabular}{|L{3.41cm}|L{5cm}|} 
			\hline 
			\textbf{Notation} & \textbf{Definition}  \\ 
			\hline 
			\( S \)					       		& Data stream \\
			\hline 
			\( I=\{I_1, I_2, ..., I_n\} \) 		& Instance window, \(I_i; (1 \leq i \leq n)\)\\
			\hline 
			\(I_t = x_t \in S \) 		   		& Incoming data instance in time t \\
			\hline
			\(y_t~/~y'_t \) 		       		& Vector of true/predicted class label \\
			\hline 
			\( C=\{C_1, C_2, ..., C_p\} \) 		& Set of p class labels, \(C_k; (1 \leq k \leq p)\)\\
			\hline 
			\( \xi=\{CS_1, CS_2, ..., CS_m\} \) & Ensemble of m individual classifiers, \(CS_j; (1 \leq j \leq m)\)\\
			\hline 
			\( s_{ij}=<S_{ij}^1, S_{ij}^2, ..., S_{ij}^p> \) & Score vector for \( I_i \) and \( CS_j \), \\
			&  \(S_{ij}^k; (1 \leq k \leq p)\)\\
			\hline 
			\( o_i = <O^1_i, O^2_i, \cdots, O^p_i> \) & Ideal-point for \( I_i \), \(O^k_i; (1 \leq k \leq p)\) \\	\hline 
			\(w = <W_1, W_2, \cdots, W_m> \) & Weight vector for \( \xi \), \(W_j; (1 \leq j \leq m)\) \\	    			 
			\hline 
		\end{tabular}
	}
\end{table}

For evaluating the performance of an algorithm in a time-evolving data stream domain, it is necessary to use tens of millions of examples \cite{bib:2009}. However, gathering this much real-world data, especially with substantial concept drifts, is not feasible. There is a shortage in trusted evolving real-world publicly available datasets for testing stream classifiers \cite{bib:surveygama2017}. Moreover, we cannot verify concept drift phases in the course of time for real-world data streams. Some popular real-world data streams, used in the literature, questionably represent sufficiently real concept drifts (e.g. discussions on electricity data \cite{bib:electcritic}). Because of these problems, like earlier studies in the literature, we use a combination of real-world and synthetic data streams in our experiments. 

We experimentally evaluate our algorithm using several real-world and synthetic datasets representing gradual, incremental, sudden/abrupt, and reoccurring concept drifts. We use the most popular real-world datasets, and for generating synthetic data streams, we use the MOA libraries \cite{bib:2009}. For the sake of comparison, we use 8 state-of-the-art ensemble methods as baselines in our experiments. We follow the tradition and use classification accuracy, processing time, and memory costs as our comparison measurements. For classification accuracy measurement, we use the Interleaved Test-Then-Train approach \cite{bib:2009}. 

\textbf{\textit{Contributions of our study.}} The summary of main contributions of this study are the following. We 

\begin{itemize}
	\item Provide a spatial modeling for online ensembles and use the linear least squares (LSQ) solution \cite{bib:lsqbook} for optimizing the weights of components of an ensemble classifier for evolving environments. While LSQ is used for batch mode component weighting \cite{bib:chan1999weighted,bib:friedman}, for the first time in the literature, we adapt and use it for online environments, as a stacking algorithm, 
	\item Introduce an ensemble algorithm, called GOOWE. We use data chunks for training, and a sliding instance window containing the latest available data for testing; such an approach provides more robust behavior, as shown by our experiments,
	\item Analyze the impact of GOOWE's weighting system on component weighting strategy and ensemble model management strategy, 
	\item  Conduct an extensive experimental evaluation on 16 synthetic and 4 real-world data streams for comparing GOOWE with 8 state-of-the-art ensemble classifiers, and 
	\item Carry out comprehensive statistical tests to show that GOOWE provides a statistically significant improvement in terms of accuracy while using conservative resources.
\end{itemize}

We present a brief chronological survey of the related work in Section 2; GOOWE in Section 3; our experimental evaluation setup in Section 4; experimental analysis in Section 5; comparative evaluation in Section 6; and statistical tests in Section 7. Section 8 offers a conclusion and directions for future research. Table \ref{tab:notation} presents the notation of symbols that we use in the succeeding sections.


\section{Background and related work}
In this section, we explain our assumptions and specifications for time-evolving data streams. We distinguish different types of concept drifts based on the literature. We discuss different approaches of adapting concept drifts in evolving environments, focusing on ensemble methods, since they are naturally more capable of handling concept drift and they proved to outperform individual classifiers \cite{bib:2009,bib:survey,bib:surveybifet2017,bib:surveygama2017}.

\subsection{Basic Concepts and Notations}
The traditional supervised classification problem aims to map a vector of attributes, \(x\), into a vector of class labels, \(y'\), i.e. \(x\mapsto y'\). The domain of attribute values in \(x\), can be either numerical or nominal. However, for the domain of class labels in \(y'\), we assume binary values for each label indicating selection or not-selection of that specific class label. We compare mapped class label vectors, \(y'\), with true class label vectors, \(y\). Instances from our data stream, 
\(
I_t = x_t \in S
\), appear sequentially in temporal order, and we must process the data in an online fashion. We map \(x_t\) into  \(y'_t\), and when the true class labels, \(y_t\), are available, we can evaluate our predictions. Due to the size of stream data, we are only able to store a limited number of instances in a window to process, and we need to discard old instances. Based on the availability of true class labels (data constraints) and our resources (solution/resource constraints), we can determine the length of the window. Classifiers are supposed to use limited memory and limited processing time per instance \cite{bib:2009,bib:survey,bib:kuncheva1}. 

In dynamically evolving environments, the conditional distribution of the output (i.e. true class labels) given the input vector, may change with time, i.e. \(P(y_{t+1}|x_{t+1}) \neq P(y_t|x_t)\), while the distribution of the input vector itself, \(P(x_t)\), may remain the same \cite{bib:survey}. This is referred to as real concept drift and has raised several challenges for detecting and reacting to these changes. 

\begin{figure}
	\centerline{\includegraphics[width=0.8\linewidth]{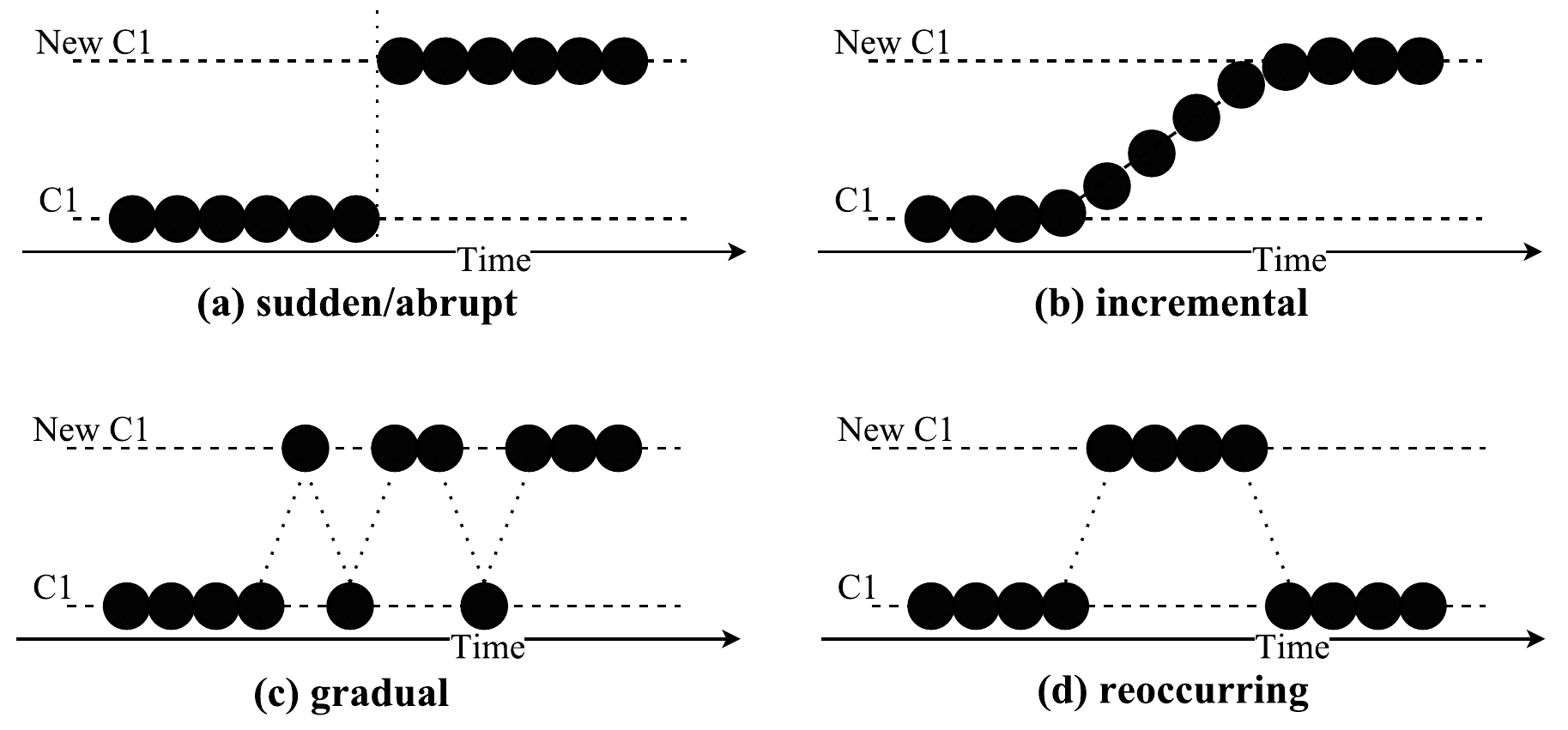}}
	\caption{ Four patterns of real concept drift over time (revised from \cite{bib:survey}).} 
	\label{fig:pattern}
\end{figure}

Zhang et al. \cite{bib:zhang} categorized real concept drifts into two scenarios;\textit{ Loose Concept Drift (LCD)} where only a change in \(P(y_t | x_t)\) causes the concept drift, and \textit{Rigorous Concept Drift (RCD)}, where change in both \(P(y_t | x_t)\) and \(P(x_t)\) cause the concept drift. The general assumption in the concept drift setting is that the change happens unexpectedly and is unpredictable. We do not consider the situation for some real-world problems where the change is predictable. We do not address concept-evolution, the arrival of a novel class label, and time-constrained classification \cite{farid2013adaptive,bib:han2015,bib:novelclass,bib:sun2016,bib:wang2015,bib:zamani}. The reader is referred to \cite{bib:survey} for various settings of the problem. We assume the most general setting of the evolving data stream classification problem.

There are several forms of change patterns over time for real concept drift, as shown in Fig. \ref{fig:pattern}. If we consider a non-changing conditional distribution of the output given the input as one concept, a drift may happen \textit{suddenly/abruptly} by replacing one concept with another (e.g. C1 with New C1 in Fig. \ref{fig:pattern}-(a)) at a moment in time t. Drift may happen \textit{incrementally} between the first and last concepts (e.g. C1 and New C1 in Fig. \ref{fig:pattern}-(b), respectively), where there are many intermediate concepts which smoothly connect the dots. \textit{Gradual drift} happens when there are no intermediate concepts and both of the first and last concepts are occurring for a period of time, Fig. \ref{fig:pattern}-(c). Drifts may introduce new concepts that were not seen before, or previously seen concepts may \textit{reoccur} after some time, Fig. \ref{fig:pattern}-(d). Once-off random anomalies or blips are called \textit{outlier/noise} and there should not be any reaction, as we do not consider them to be concept drift. Since most of the real-world problems are complex mixtures of these concept drifts, we expect any classifier to react and adapt reasonably to different types of concept drifts and remain robust to outliers, predicting with acceptable resource requirements \cite{bib:survey}. 	

\subsection{Ensemble Classifiers for Evolving Online Environments}	
A recently published survey on concept drift adaption \cite{bib:survey}, presents a new taxonomy of adaptive classifiers using four existing modules of various learning methods in time-evolving environments. They are \textit{memory management}, \textit{change detection}, \textit{learning property}, and \textit{loss estimation}. In this study, we concentrate on model management strategies, as a learning property, to present state-of-the-art ensemble methods in chronological order. Model management strategies are techniques used in maintaining ensemble components as new data become available in the course of time. In addition, since we provide a novel stacking algorithm for online ensemble classifiers, we cover vote combination techniques of these ensembles. The remaining modules, other than learning property, are out of the scope of this paper.   

Two more recently published surveys on ensemble learning for data stream analysis \cite{bib:surveybifet2017,bib:surveygama2017} show the importance of ensemble learning methods, especially on changing environments, and present ongoing research challenges. Gomes et al. cover existing data stream ensemble learning methods, propose a consistent taxonomy among them, and compare them based on some important aspects like vote aggregation, diversity measurement, and dynamic updates \cite{bib:surveybifet2017}. Krawczyk et al. discuss more advanced topics such as imbalanced data streams, novelty detection, active and semi-supervised learning, complex data representations, and structured outputs with a focus on ensemble learning \cite{bib:surveygama2017}.

Based on the model management categories of Kuncheva \cite{bib:kuncheva1}, there are five possible strategies for adaptive online classifiers: 

\begin{enumerate}
	\item \textit{Horse Racing:} The dynamic combination ensemble strategy that aims to have the most proper combination rule of existing individual components in an ensemble; 
	\item \textit{Updated Data Feeding:} Feeding individual classifiers with the most recent available data; 
	\item \textit{Scheduled Feeding of Ensemble Members:} Scheduling the update of individual classifiers, either by retraining in a batch mode, or incrementally in an online mode with newly available data;
	\item \textit{Add/Drop Classifiers:} Adding fresh classifiers to the ensemble or pruning the deteriorating classifiers; and
	\item \textit{Feature Regulation:} Regulating the importance of features along with the life of an ensemble.
\end{enumerate}
 Practically any combination of these strategies can be used together and they do not need to be necessarily mutually exclusive.

Elwell and Polikar \cite{bib:nse} explain \textit{active} versus \textit{passive} approaches. Active approaches benefit from a drift detection mechanism, reacting only when drift is detected. On the other hand, passive approaches continuously update the model with each incoming data. Since training identical hypotheses with the same data produces identical classifiers, we need some mechanisms to increase their diversity. This is accomplished mostly by Kuncheva's third and fourth strategies. In addition, there are some works to measure and maintain the diversity of component classifiers \cite{bib:minku1,bib:minku2}.

The WINNOW \cite{bib:winnow}, Weighted Majority (WM) \cite{bib:wm}, and Hedge(\( \beta \)) \cite{bib:hedge} algorithms are the initial adaptive ensemble methods for large-scale changing environments. They mainly use the horse racing strategy for developing better combination rules in an off-line setting. They begin by creating a set of classifiers with an initial weight (usually 1). They adapt the ensemble's behavior using a reward-punishment system to keep track of the most trustworthy expert in each time slot. In particular, WINNOW uses \(\alpha>1\) (usually \(\alpha=2\)) for its promotion (\(w_i \leftarrow w_i \times \alpha \)) and demotion (\(w_i \leftarrow w_i \div \alpha \)) steps. WM excludes the promotion step, and if an expert incorrectly classifies the instance, the algorithm decreases its weight by a multiplicative constant, \(\beta \in [0, 1]\). The Hedge(\( \beta \)) algorithm operates in the same way, but instead of taking the weighted majority vote, chooses one classifier's decision as the ensemble decision. They provide a general framework for weighting component classifiers. However, they do not suggest any mechanism for dynamically adding or removing new components.

The Streaming Ensemble Algorithm (SEA) \cite{bib:sea} provides a block-based and fixed-size ensemble of classifiers, each trained on the incoming chunk of instances---addressing Kuncheva's fourth model management strategy. If the ensemble has space, SEA adds the new classifier to the ensemble; otherwise, it puts the new classifier in the place of a weaker classifier. SEA uses a majority vote for predictions in an off-line setting. Due to batch mode component classifiers stopping learning after being formed, replacing the worst performing classifier in an unweighted ensemble, the learner is unable to properly track concept drifts in the stream data.

Oza \cite{bib:ozaphd,bib:oza2} uses Kuncheva's second and third model management strategies together with the traditional bagging and boosting algorithms in online settings for designing  OzaBagging and OzaBoosting. For stream data environments, as the number of training examples and component classifiers tend to go to infinity, Oza uses the Poisson distribution with \( \lambda = 1\) for approximating the binomial distribution of sampling. A similar idea is used for the OzaBoosting algorithm. It employs incremental values of \(\lambda\), starting from 1, for the training and sampling of classifiers. 

The Dynamic Weighted Majority (DWM) \cite{bib:dwm2,bib:dwm} introduced an ensemble of incremental learning algorithms, each with an associated weight in an online setting. Models are generated by the same learning algorithm on different batches of data. DWM uses the WM approach for assigning weights and makes predictions using a weighted-majority vote of the components where weights are dynamically changing. Pruning components with weights less than a threshold helps to avoid creating an excessive number of components. An extension to DWM, additive expert ensemble (AddExp) \cite{bib:addExpert}, provides a general theoretical expert analysis to prove mistakes and loss bounds for a discrete and a continuous ensemble. 

The Accuracy Weighted Ensemble (AWE) \cite{bib:awe} alternatively suggests a general framework for mining changing data streams using weighted ensemble classifiers by re-evaluating ensemble components with incoming data chunks. Inspired by the framework of SEA, a new static learning algorithm is trained and the previous components of ensemble are evaluated on each incoming data chunk. However, these evaluations are done with a special version of Mean Square Error (MSE) allowing the algorithm to select the \(k\) best classifiers to create a new ensemble (\(MSE_i=\frac{1}{|D|}\sum_{x \in D}(1-M^i_c (x))^2\); where \(D\) is the latest data chunk and \( M^i_c (x)\) is the probability score that \(x\) belongs to its true class label \(c\), generated by a specific classifier system indexed \(i\)). Briefly, it assigns weights to component classifiers based on their expected classification accuracy---according to Bayes error optimization \cite{bib:decbound}. Moreover, the structure of the ensemble is pruned if errors of individual classifiers are worse than the MSE of a random classifier (\( MSE_r = \sum_{c} P(c) \times (1-P(c))^2\); where \(P(c)\) is the probability of observing class label \(c\)). All in all, the weight of classifier \(i\) is determined by a linear function (\( w_i = MSE_r - MSE_i \)).

\begin{table*}
	\tbl{Summary of Related Ensemble Classifiers for Evolving Online Environments\label{table:related}}
	{	
		\begin{tabular}{lllccccc}
			\toprule
			& \multicolumn{2}{c}{Spec.} & \multicolumn{5}{c}{Kuncheva's Strategies} \\
			\cmidrule(lr){2-3}\cmidrule(lr){4-8}
			Ensemble         & Study & Type & St. 1 & St. 2 & St. 3 & St. 4 & St. 5 \\
			\midrule
			WINNOW            & \cite{bib:winnow}  &  Passive  & \checkmark & $\times$ & \checkmark & $\times$  & $\times$  \\
			WM                & \cite{bib:wm}  &  Passive  & \checkmark & $\times$ & \checkmark & $\times$  & $\times$  \\
			Hedge(\(\beta\))  & \cite{bib:hedge}  &  Passive  & \checkmark & $\times$ & \checkmark & $\times$  & $\times$   \\
			SEA               & \cite{bib:sea}  &  Passive  & $\times$ & $\times$ & \checkmark & \checkmark  & $\times$   \\
			OzaBag/OzaBoost   & \cite{bib:ozaphd,bib:oza2}  &  Passive  & $\times$ & \checkmark & \checkmark & $\times$  & $\times$   \\
			DWM               & \cite{bib:dwm2,bib:dwm}  &  Passive  & \checkmark & $\times$ & \checkmark & \checkmark  & $\times$   \\
			AWE               & \cite{bib:awe}  &  Passive  & \checkmark & $\times$ & \checkmark & \checkmark  & $\times$   \\
			ACE               & \cite{bib:ace}  &  Active  & \checkmark & \checkmark & \checkmark & $\times$  & $\times$   \\
			LevBag            & \cite{bib:lev}  &  Active  & \checkmark & \checkmark & \checkmark & \checkmark  & $\times$   \\
			Learn++.NSE       & \cite{bib:nse}  &  Passive  & \checkmark & $\times$ & \checkmark & \checkmark  & $\times$   \\
			AUE2              & \cite{bib:aue}  &  Passive  & \checkmark & $\times$ & \checkmark & \checkmark  & $\times$   \\
			OAUE              & \cite{bib:oaue}  &  Passive  & \checkmark & \checkmark & \checkmark & \checkmark  & $\times$   \\
			\midrule
			GOOWE              & Current work  &  Passive  & \checkmark & $\times$ & \checkmark & \checkmark  & $\times$   \\
			\bottomrule	
		\end{tabular}
	}
\end{table*}

Since larger data chunks can provide a better distribution of data, they are more capable of building more accurate classifiers but may contain more than one change. Smaller chunks can separate drifting places better, but usually lead to poorer classifiers. In particular, ensembles built on large data chunks may react too slowly to sudden drifts occurring inside the chunk \cite{bib:2009,bib:aue}. To overcome this problem, Adaptive Classifier Ensemble (ACE) \cite{bib:ace}, proposed an algorithm which uses a hybrid of one online classifier and a collection of batch classifiers (a mixture of active and passive approaches) along with a drift detection mechanism. ACE does not benefit from pruning strategies, and the possible use of a drift detector leads to poor reactions for gradual drifts.

Bifet \cite{bib:lev} introduced Leverage Bagging (LevBag) as an extended version of OzaBagging, using the first four strategies of Kuncheva. It aims to increase the resampling rate using a larger value of \(\lambda\) in the Poisson distribution. Additionally, it adapts output detection codes \cite{bib:output} for handling multi-class problems using only binary classifiers and the ADWIN  \cite{bib:adwin} change detector for dealing better with concept drifts in stream data. 

Learn++.NSE (NSE) \cite{bib:nse} is a batch learning ensemble that uses weighted majority voting. It updates weights dynamically with respect to the time-adjusted errors of the classifiers on current and past environments. Similar to the AWE model management approach, evaluation of classifiers is considered by giving more credit to the ones capable of identifying previously unknown instances. On the other hand, classifiers that misclassify previously known instances are penalized. Moreover, NSE does not discard any component from the ensemble when its knowledge is not relevant to the current chunk of data. Although temporarily forgetting model management is particularly useful in cyclical environments, it causes some resource overuse. Ditzler and Polikar extended NSE for class imbalanced data stream \cite{bib:nseimb}.  

Brzezinski et al. \cite{bib:aue} proposed the Accuracy Updated Ensemble (AUE2), for combining the chunk-based algorithms with incremental learning components. Its model management strategy is based on AWE, and suggests a non-linear weighting function using the same MSE functions (\(w_{ij}=\frac{1}{(MSE_r+MSE_{ij}+\epsilon)} \)). The online version of AUE2 \cite{bib:oaue}, called Online Accuracy Updated Ensemble (OAUE), uses a sliding window for the last $n$ instances of the data stream.

A summary of these online ensemble classifiers is provided in Table \ref{table:related}. Our ensemble, GOOWE, that we present in the next section, is also included in the table for comparison. As we can see, GOOWE's model management strategies are the same as AWE and AUE2.

\textit{\textbf{Ensemble size.}} It is also called ensemble cardinality in some studies. Determining the number of component classifiers for an ensemble, discussed briefly in \cite{bib:surveybifet2017,bib:surveygama2017}, is an important problem since it has high impact on the prediction ability of an ensemble, and resource consumptions, in terms of time and memory. Our study \cite{bib:bonabcikm} shows that the intuition of adding more classifiers will result in greater accuracy, is incorrect in practice. In the context of data stream classification, the ensemble size can either be defined fixed, or dynamic, prior to the execution. 
While there is a lack of studies for determining the size of an online ensemble, most of the existing studies for batch ensembles use statistical tests for determining the proper number of components \cite{bib:latinne2001limiting,bib:oshiro2012many,bib:hernandez2013large}. Our geometric framework used for the weighting of components of GOOWE, is also used for determining the ideal number of classifiers for online ensembles, in a theoretical perspective. Increasing or decreasing the number of classifiers from this ideal point deteriorates predictions. We called it ``the law of diminishing returns in ensemble construction.'' Our theoretical study shows that using the same number of independent component classifiers as class labels gives the highest accuracy \cite{bib:bonabcikm}.

\section{GOOWE: Geometrically Optimum and Online-Weighted Ensemble}
\textbf{\textit{Concepts and Motivation.}} Unlike traditional batch learning, the assumption of independent and identical distribution (i.i.d) of the whole stream data is not true for evolving online environments \cite{bib:gama2013evaluating}. The possibilities of changes are; ``feature changes'', or evolving of $p(x)$ with time stamp $t$, ``conditional change'',  or the changes of class label $y$ assignment to feature vector $x$, and ``dual changes'', which includes both \cite{bib:gao}. Four recognized patterns of conditional change are given in Fig. \ref{fig:pattern}. The same patterns of change are possible for feature changes. As mentioned in Section 2.1, Zhang et al. \cite{bib:zhang} categorized these change into LCD and RCD scenarios. An effective classification algorithm should be able to handle these continuous changes. 

The data stream is sliced into chunks, each representing a single distribution. Almost all state-of-the-art stream classifiers divide the data into fixed chunk sizes, as $h$ \cite{bib:mustafa}. There is a recent study for dynamic determination of chunk size according to concept drift speed \cite{bib:mustafa}. This problem is beyond the scope of our study. 

Depending on when the labeled training data becomes available, Gao et al. \cite{bib:gao} categorized stream classifiers into two groups: The first group updates the training distribution as soon as the labeled instance becomes available, and the second group receives labeled data in chunks and updates the model. Since updating classifiers is a costly operation, the second group of classifiers can be more time efficient. However, these methods perform well when the up-to-date data chunk has identical or similar distributions to the yet-to-come data chunk, which is called a stationary assumption in the data stream. This assumption ignores the instable nature of evolving data streams when concept drift occurs frequently. 

\begin{figure}
	\centerline{\includegraphics[width=\linewidth]{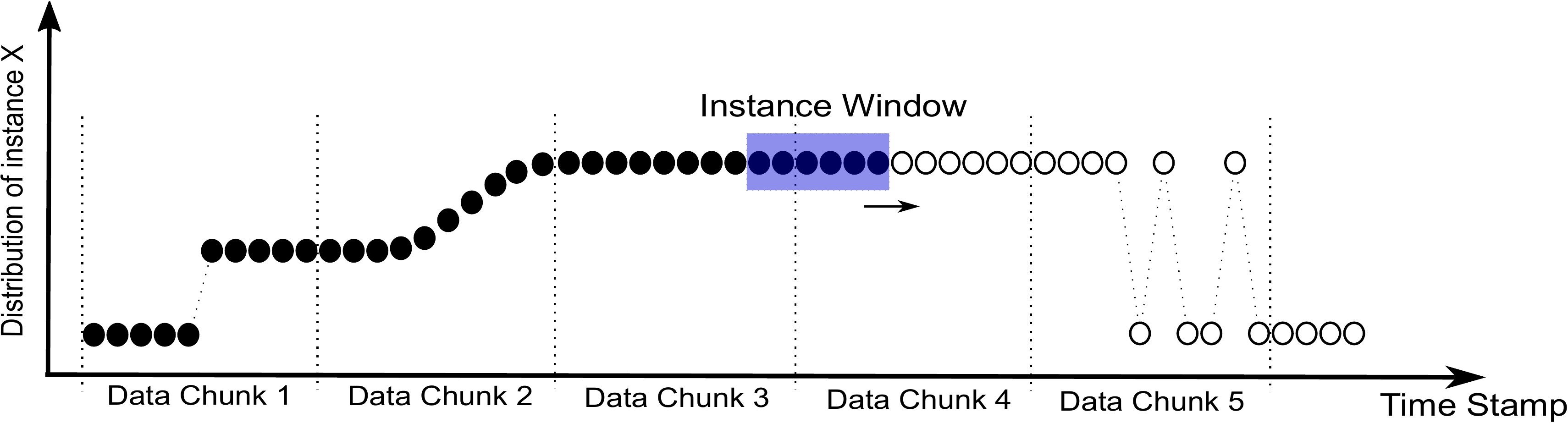}}
	\caption{Data Chunk (\(DC\) ) vs. Instance Window (\(I\))---stream data is sliced into equal chunks with size of $h$ and sliding instance window takes the latest $n$ instances with available labels; filled circles are instances with available labels and unfilled circles are yet-to-come instances. } 
	\label{fig:chunk}
\end{figure}

To make our ensemble more efficient, we update component classifiers when a new chunk of labeled data is received. Although we do not address concept drift adaption directly, our extensive experiments show that using a proper component weighting system based on very recent instances would adapt existing component classifiers for recent concept changes. Consequently, having an optimum weighting function would be extremely beneficial for handling concept drift. For this purpose, we exploit a sliding instance window with the latest $n$ labeled instances. The size of the instance window can vary with chunk size, $h \neq n$. The instance window size can be determined by the performance and accuracy requirements of the problem. Fig. \ref{fig:chunk} shows this combination usage of data chunk and instance window. 

Inspired from the geometry of voting \cite{bib:saari2008} and using the least squares problem (LSQ) \cite{bib:lsqbook}, we designed a geometrically optimum and online-weighted ensemble method for evolving environments, called GOOWE. While LSQ is used for component weighting of ensemble classifiers in batch mode \cite{bib:chan1999weighted,bib:friedman}, it is the first time that we provide a spatial modeling for online environments as a stacking algorithm.

The motivation of this study is to design an ensemble that assigns optimum weights to component classifiers, in an on-line setting with different types of concept drifts. For combining votes, as a stacking algorithm, we model scores of the ensemble's individual classifiers in a spatial environment as vectors, and try to establish a clear relationship between a geometric feature of vectors, and their effectiveness. Its novelty is based on a dynamically changing component optimum weight assignment approach for online ensembles in evolving data streams.

\textbf{\textit{Design.}} GOOWE's model management approach is similar to AWE and AUE2, with a passive approach for handling concept drift. Basically, a new incremental learning algorithm is trained on each incoming data chunk, and the previous components of the ensemble are re-evaluated on the same data chunk. However, these evaluations are done with a special function of mean square error (MSE), allowing the algorithm to assign the weights of component classifiers dynamically, relative to each other, and in an on-line setting. 

\begin{figure}
	\centerline{\includegraphics[width=0.9\linewidth]{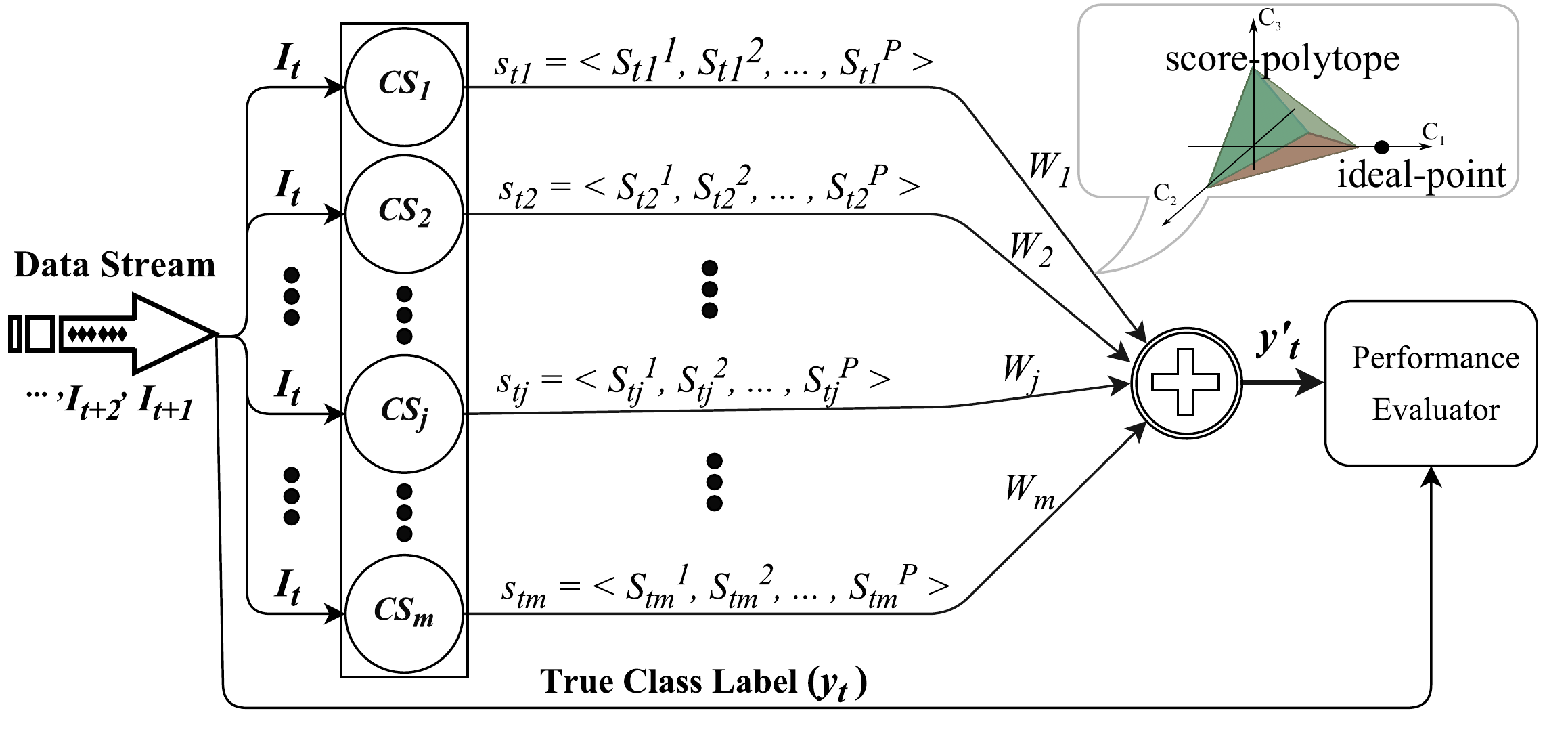}}
	\caption{General schema of GOOWE; each \(I_t \in S\) delivered to \(CS_j (1 \leq j \leq m)\), produces relevance score vector, \( s_{tj} \). GOOWE maps these score vectors, as a score-polytope, and the true class label, as an ideal-point, to a p-dimensional space. It assigns weights, \(W_j\), using the linear least squares (LSQ) solution. The predicted class label, \(y'_t\), is obtained using the weighted majority voting \cite{bib:bonabcikm}. } 
	\label{fig:ensemble}
\end{figure}

In the training scenario, we use data chunks according to Fig. \ref{fig:chunk}, as they become available. When a new data chunk is received, we train a new component classifier using these instances and we add it to the ensemble. If there is no space for the new classifier, we substitute it with the worst-performing component. For testing the ensemble and classifying a new instance, we use our LSQ-based stacking algorithm based on the sliding instance window for getting the most updated weights for adapting existing components. Briefly, GOOWE uses a combination of data chunks and instance windows, as shown in Fig \ref{fig:chunk}. A \textit{data chunk (DC)} has \(h\) instances of a equally divided data stream; an \textit{instance window (I)} has the latest \(n\) instances of a data stream, with available true class labels. In our implementation, we build the instance window with the length of \( max(n,h)\), and simply add a counter with the maximum value of \(h\) into the instance window for providing the data chunk. If the length of the instance window is less than the length of data chunk (i.e. \(n<h\)), we set the length of instance window to \(h\) and use the latest \(n\) instances.    

In our geometric framework, we use the Euclidean norm as the system's loss function for optimization purposes. There are clear statistical, mathematical, and computational advantages of using the Euclidean norm \cite{bib:lsqbook}. We calculate weights based on the latest \(n\) instances in our window, and to make a prediction we use a weighted majority voting approach.

As shown in Fig. \ref{fig:ensemble}, we have an ensemble of component classifiers \(\xi = \{CS_1, CS_2, \cdots , CS_m\}\). Each component classifier, \(CS_j (1 \leq j \leq m) \), processes instance \(I_t\) of an evolving data stream, \(S\), and produces relevance scores, \(s_j=<S_j^1, S_j^2, \cdots, S_j^p>\), for each of the class labels, \(C = \{C_1, C_2, \cdots, C_p\}\). Since each classifier produces relevance scores in different ranges, we use Eq. \ref{eq:normal1} for normalizing the scores into the range of [0, 1]. 

\begin{equation}\label{eq:normal1}
	S_j^k \leftarrow \frac{S_j^k}{\sum_{a=1}^{p} S_j^a} ~~ (1 \leq k \leq p)
\end{equation}  

Taking each class label as one dimension, enables us to map each component's score \((s_j; 1 \leq j \leq m)\) into a point in a p-dimensional Euclidean space. Mapping all score points of \(I_t\) in the same way, builds a polytope in a p-dimensional Euclidean space, which we call the \textit{score-polytope} of \(I_t\). We define \textit{score-vector} by using the origin point as the starting point and score point as the terminal point in our spatial environment. Using the vector of the true class label for \(I_t\) as \(y_t\), we can assume an \textit{ideal-point} in the p-dimensional space as \(o = <O^1, O^2, \cdots, O^p>\). For example, if the number of class labels is 4, and the true class label of \(I_t\) is \(C_2\), then the ideal-point would be \(o = <0, 1, 0, 0>\). 

\textbf{\textit{Optimum Weight Assignment.}} For making predictions, we use \(n\) latest instances \(I = \{I_1, I_2, \cdots, I_n\} \), as an instance window, where \(I_n\) is the latest instance and all true class labels are available. For each instance \(I_i (1 \leq i \leq n)\), each component classifier \( CS_j (1 \leq j \leq m) \) has a score-vector as \( s_{ij} = < S^1_{ij}, S^2_{ij}, \cdots, S^p_{ij} >\). For the true class label of \(I_i\) we have \( o_i = <O^1_i, O^2_i, \cdots, O^p_i> \) as the ideal-point. We aim to find the optimum weight vector \(w = <W_1, W_2, \cdots, W_m> \) to minimize the distance between score-polytope and ideal-point. Using the squared Euclidean norm as our measure of closeness for the linear least squares problem (LSQ) results in
\begin{equation}
	\min_w || o - Sw ||^2_2
\end{equation} 
The corresponding residual vector is \(r=o-Sw\), where for each instance \( I_i\), \( S \in \mathbb{R}^{m \times p} \) is the matrix with relevance scores \(s_{ij}\) in each row, \(w\) is the vector of weights to be determined, and \(o\) is the vector of ideal-point \cite{bib:lsqbook}. Since we have \(n\) instances in our window, we use the following function for our optimization solution.  
\begin{equation}
	f(W_1, W_2, \cdots, W_m) = \sum_{i=1}^{n} \sum_{k=1}^{p} (\sum_{j=1}^{m} (W_j S^k_{ij}) - O^k_i)^2
\end{equation}
Taking a partial derivation over \(W_q (1 \leq q \leq m)\) and finding optimum points will give us our weight vector. The gradient equations become
\begin{equation}
	\frac{\partial f}{\partial W_q} = \sum_{i=1}^{n} \sum_{k=1}^{p} 2(\sum_{j=1}^{m} (W_j S^k_{ij}) - O^k_i)S^k_{iq}  ,~~(1 \leq q \leq m)
\end{equation}
Setting the gradient to zero, \( \nabla f = 0 \) 
\begin{equation}
	\sum_{j=1}^{m} W_j(\sum_{i=1}^{n} \sum_{k=1}^{p} S^k_{iq} S^k_{ij})
	= \sum_{i=1}^{n} \sum_{k=1}^{p} O^k_i S^k_{iq} ,~~(1 \leq q \leq m)
\end{equation}
and assuming below summations as \(a_{qj}\) and \(d_q\) 
\begin{equation} \label{eq:aqj1}
	a_{qj} = \sum_{i=1}^{n} \sum_{k=1}^{p} S^k_{iq} S^k_{ij} ,~~(1 \leq q, j \leq m)
\end{equation}
\begin{equation}\label{eq:dq1}
	d_q = \sum_{i=1}^{n} \sum_{k=1}^{p} O^k_i S^k_{iq} ,~~(1 \leq q \leq m)
\end{equation}
lead to \(m\) linear equations with \(m\) variables (weights). The proper weights in the following matrix equation are our intended optimum weight vector. We present the weight assignment equation in matrix representation to make the later example easier to follow.
\begin{equation} \label{eq:linsys1}
	\begin{bmatrix}
		a_{11} & a_{12} & \cdots & a_{1m}           \\[0.3em]
		a_{21} & a_{22} & \cdots & a_{2m}           \\[0.3em]
		\vdots & \vdots & \ddots & \vdots           \\[0.3em]
		a_{m1} & a_{m2} & \cdots & a_{mm}           \\[0.3em]
	\end{bmatrix} \times
	\begin{bmatrix}
		W_1 \\[0.3em]
		W_2  \\[0.3em]
		\vdots \\[0.3em]
		W_m  \\[0.3em]
	\end{bmatrix} = 
	\begin{bmatrix}
		d_1 \\[0.3em]
		d_2  \\[0.3em]
		\vdots \\[0.3em]
		d_m  \\[0.3em]
	\end{bmatrix} 		
\end{equation}
Briefly, \(	Aw=d \), where \( A \) is the coefficients matrix and \(d\) is the remainders vector. According to Eq. \ref{eq:aqj1}, \( A \) is a symmetric square matrix. In the sense of the least squares solution \cite{bib:lsqbook}, since it is probable that \( A \) is rank-deficient, we may not have a unique solution and we denote the minimizer by \( w^* \). According to Theorem 9 of \cite{bib:lsqbook}, the \textit{normal equations} for \( w^* \) can be written as  
\begin{equation}
	A^T A w = A^T d
\end{equation}
In this equation, \( A^TA \), is also a symmetric square matrix. In addition, if \( A \) has full rank, \( A^TA \) is positive definite and our problem has a unique solution. In the rank-deficient case, it is a non-negative definite, and we have a set of possible weight vectors. The QR factorization suggests less expensive solutions for both full rank and rank-deficient cases \cite{bib:lsqbook}. In such cases, the weights are nearly optimal.

Since we predict scores for each incoming instance separately, we define \( A_i \) and \(d_i (1 \leq i \leq n)\) according to Eq. 6 and Eq. 7. Matrix \( A \) and vector \(d\) can be calculated simply by adding all \( A_i \) and \(d_i\) for all instances of a given window, respectively. 
\begin{equation} \label{eq:aqj2}
	a_{qj}^i = \sum_{k=1}^{p} S^k_{iq} S^k_{ij},~~(1 \leq i \leq n)
\end{equation}
\begin{equation} \label{eq:dq2}
	d_q^i = \sum_{k=1}^{p} O^k_i S^k_{iq},~~(1 \leq i \leq n)
\end{equation}

Using the weighted majority vote approach gives the aggregated score vector. Since we calculate scores in a spatial environment, it is possible that these score values become negative. Using the following normalization before Eq. \ref{eq:normal1} gives the proper aggregated score vector.

\begin{equation}\label{eq:normal2}
	S_k \leftarrow \frac{S_k - min(S_k)}{ max(S_k) - min(S_k) } ,~~ (1 \leq k \leq p)
\end{equation}

\begin{figure}[h]
	\centerline{\includegraphics[width=0.7\linewidth]{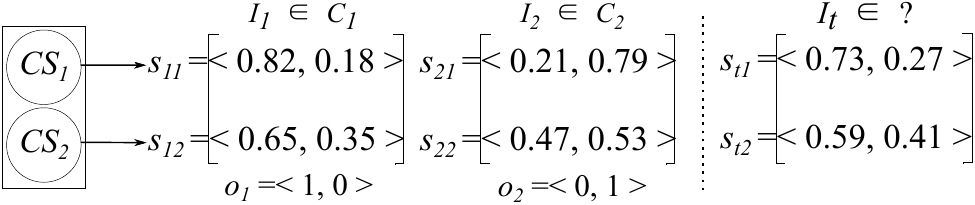}}
	\caption{An example of GOOWE component classifiers weighting.} 
	\label{fig:gwmyfig}
\end{figure}

\textbf{\textit{Example - Assigning Optimal Weights for Component Classifiers.}} Suppose that we have 2 classifiers and 2 class labels, as shown in Fig. \ref{fig:gwmyfig}. Our instance window has 2 instances as \( I_1\) and \( I_2\). We want to find the optimum weight vector for aggregating scores for a newly arrived instance as \( I_t\).

\begin{figure}[h]
	\centerline{\includegraphics[width=0.8\linewidth]{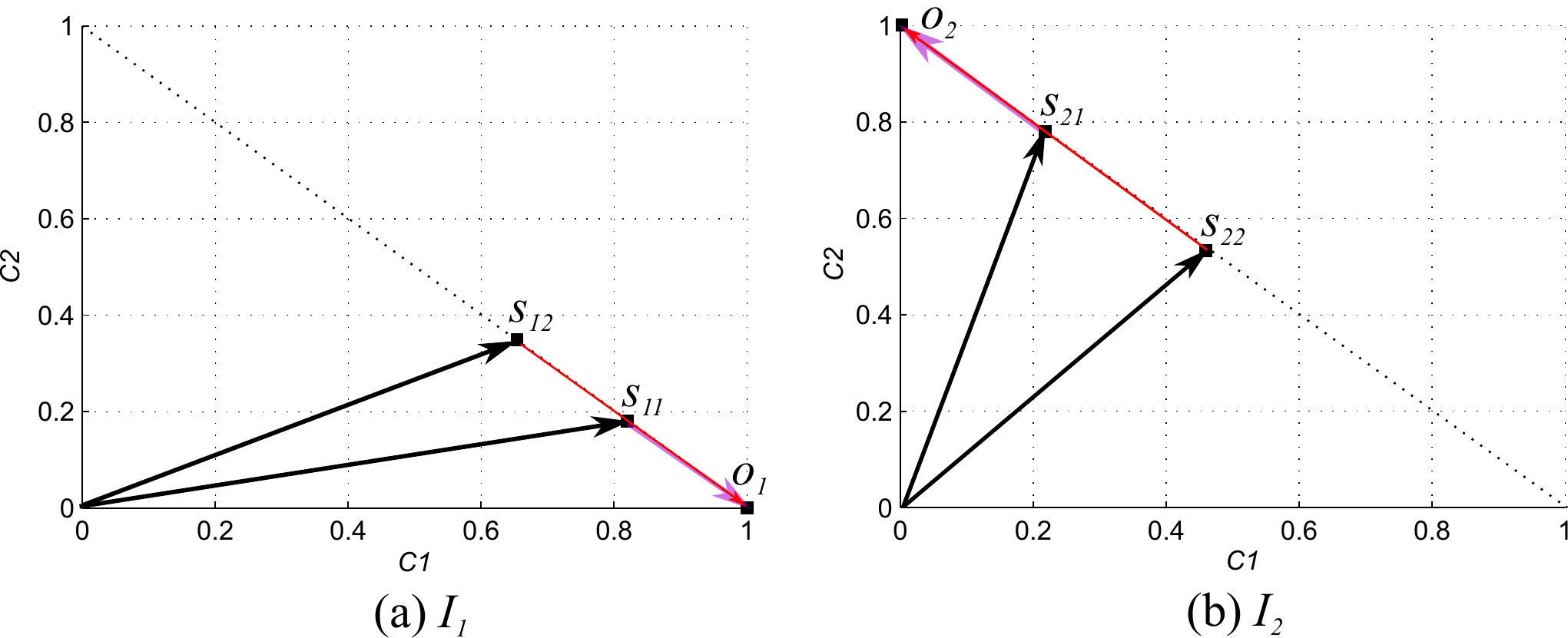}}
	\caption{Score vectors for instance window of example.} 
	\label{fig:vector}
\end{figure}

We have a 2-dimensional Euclidean space, as shown in Fig. \ref{fig:vector}. Score vectors and their intended projections are illustrated with black and red lines, respectively. Putting the values into Equation \ref{eq:aqj1} and \ref{eq:dq1}, gives the following matrix equation.  

\[
\begin{bmatrix}
1.37 & 1.11           \\[0.3em]
1.11 & 1.05           \\[0.3em]	  
\end{bmatrix} \times
\begin{bmatrix}
W_1 \\[0.3em]
W_2  \\[0.3em]      
\end{bmatrix} = 
\begin{bmatrix}
1.61 \\[0.3em]
1.18  \\[0.3em] 
\end{bmatrix} 		
\]

Solving this equation gives the intended weight vector, \( w = <1.88, -0.87> \). Multiplying these weights with the score vectors of the components, results in the aggregated score vector, \( s = < 0.86, 0.14 > \). We have a much stronger vote compared to each individual classifier.  

\begin{algorithm*}
	
	\caption{GOOWE (Geometrically Optimum and Online-Weighted Ensemble) }
	
	\begin{algorithmic} [1]
		\REQUIRE {$S\colon$data stream, $I\colon$window of \textit{n} latest instances, 
			$DC\colon$latest data chunk with length of \textit{h}, $m\colon$maximum number of classifiers, $~CS\colon$single classifier system, $~p\colon$number of class labels, $~L\colon$memory limit. }
		\ENSURE {$\xi\colon$set of weighted classifiers, $s_T\colon$aggregated score vector. }
		
		\STATE $\xi \leftarrow \emptyset$;		
		\WHILE{$S$ has more instances}											
		\FORALL{instances $I_i \in I$ }				
		\STATE $A \leftarrow A + A_i$;
		\COMMENT{using Eq. \ref{eq:aqj2}}
		\STATE $d \leftarrow d + d_i$;
		\COMMENT{using Eq. \ref{eq:dq2}}
		\ENDFOR 
		\STATE $w \leftarrow solve(Aw=d)$;
		\COMMENT{see Eq. \ref{eq:linsys1}}
		\STATE $s_T \leftarrow \sum_{j=1}^{m} (W_j s_j$);
		\COMMENT{weighted majority vote}
		
		\IF{$DC~has~h~instances$} 			
		\STATE $CS' \leftarrow$ new single classifier built on DC;
		\IF{$\xi~has~m~classifiers $}  
		\FORALL{instances $I_i \in DC$ }				
		\STATE $A' \leftarrow A' + A'_i$;
		\COMMENT{using Eq. \ref{eq:aqj2}}
		\STATE $d' \leftarrow d' + d'_i$;
		\COMMENT{using Eq. \ref{eq:dq2}}
		\ENDFOR 
		\STATE $w' \leftarrow solve(A'w'=d')$;
		\COMMENT{see Eq. \ref{eq:linsys1}}
		\STATE $\xi \leftarrow \xi \setminus $\{classifier with $min(|W'_j|)$; $ 1\leq j \leq m $\}; 
		\ENDIF
		\FORALL{ $CS_j \in \xi$ }				
		\STATE train $CS_j$ with $DC$;						
		\ENDFOR 
		\STATE $\xi \leftarrow \xi \cup \{CS'\}$;
		\ENDIF
		
		\IF{$memory\_usage(\xi) \geq L $}  
		\STATE $prune$ all component classifiers;					
		\ENDIF
		
		\ENDWHILE 
		
	\end{algorithmic}
	
\end{algorithm*}

\textbf{\textit{Pseudocode of GOOWE Algorithm.}} It is given in Algorithm 1. In the training scenario (lines 9-23), having the proper number of instances from each class label, as our training data, is crucial for more accurate individual classifiers. On the other hand, for the testing scenario (lines 3-8), static weighting component classifiers can result in relatively poor aggregated predictions, especially with the existence of frequent concept drifts in data stream. Using a combination of data chunk and instance window enables us to think about training and testing of our algorithm separately. These two values can be adjusted according to the drift rate of the data stream.

When the number of instances in the data chunk, DC, reaches its maximum value (line 9), GOOWE trains a new incremental classifier (line 10). If the ensemble has its maximum number of classifiers, \(m\), then GOOWE calculates the weights of classifiers using Eq. \ref{eq:linsys1} and the instances in the data chunk (lines 12-16). The more the obtained weight value is close to zero, the more we want to cancel its effectiveness in our aggregated score vector. As a result, we take the absolute value of weight values and omit the classifier with the least weight (line 17). We first incrementally update all the existing classifiers with DC (lines 19-21), and then add a fresh classifier into the ensemble (line 22). Most of the incrementally updated classifiers need to be pruned after some updates. Since we have memory constraints in our problem, we prune these classifiers when the consumed memory exceeds the memory limit (lines 24-26). For example, in our experiments we use the Hoeffding tree \cite{bib:ht}, and prune the least active leaves of the tree to satisfy the user-specified memory constraint.

For making the class label prediction for each incoming instance, GOOWE calculates the weights of classifiers using Eq. \ref{eq:linsys1} and the instances in the instance window (lines 3-7). It multiplies the resulting weights with score vectors, and using the weighted majority voting approach, calculates the aggregated score vector. Adjusting the length of the instance window and data chunk depends on the data stream and types of concept drift. There is no general solution to this problem. However, setting relatively small values to the instance window, and relatively large values to the data chunk, according to available resources, can result in better accuracy. 

Experimental evaluations, presented in the following sections, illustrate that GOOWE can react statistically significantly better compared to its state-of-the-art rivals.   

\section{Experimental Evaluation Setup}
The main concerns for evolving data stream classifiers are more accurate predictions with less memory consumption, and less processing time. In addition, any proposed method for an evolving data stream needs to be careful with concept drift, and react accordingly. In the following sections, we present our experimental evaluation for different simulation scenarios conducted to evaluate our proposed ensemble. 

In summary, our experimental evaluation is presented as follows. We 
\begin{itemize}
	\item \textit{First,} describe synthetic and real-world evolving data streams used in our experiments. We explain each with the type of concept drift, the number of class labels, and the number of instances used. While there is a shortage in trusted evolving real-world streams \cite{bib:surveygama2017}, we try to include all possible known/unknown categories of concept drift in our experiments. We also specify our experimental framework setup, implementation details, used libraries, and more for reproducibility purpose (current section). 
	\item \textit{Second,} provide an analysis conducted for examining two major differentiating elements of GOOWE, component weighting strategy and ensemble model management strategy (Section 5). Our optimum and online weighting system shows its effectiveness for both vote aggregation and ensemble maintenance. 
	\item \textit{Last but not least,} present our extensive and comparative experiments. We compare GOOWE with state-of-the-art rival ensembles and extensively discuss the superiority conditions. For the sake of comparison, we include 8 state-of-the-art adaptive ensemble methods proposed for evolving data streams (Section 6). 
\end{itemize}

\subsection{Datasets as Data Streams}
Selecting proper time-evolving data streams is one of the vital steps for comparing different algorithms. There are two types of data stream sets---synthetic and real-world datasets. We generate the whole dataset before the experiment, and use the terms dataset  and data stream equivalently. Similarly to other domains of prediction algorithms, real-world datasets are the best. However, their problem is that we do not know when drift occurs, or if there is any drift at all. Some studies use real-world datasets with artificial concept drifts, called real-world data with forced/synthetic concept drift \cite{bib:survey}. These datasets cannot be considered as real examples of drifts. Synthetic data has several benefits like being easy to reproduce, having a low cost of storage and transmission, but most importantly, it provides an advantage of knowing where exactly drift has happened \cite{bib:2009,bib:survey}. 

A proposed algorithm should be capable of handling large data streams---with potentially an infinite number of instances \cite{bib:2009}. As a result, for the comparison of several algorithms, we need to have large datasets in the order of tens of millions of instances. Similar to common approaches \cite{bib:2009,bib:aue,bib:oaue,bib:sea}, in order to cover all patterns of changes over time; sudden/abrupt, incremental, gradual, and reoccurring as concept drifts including blips or noise; we use synthetic data stream generators, implemented in the MOA framework. Using these generators, we prepared 16 synthetic datasets. In addition, we have 4 widely used real-world data streams.

Following are a brief description of each dataset including their generation and preparation. Table \ref{tab:Dataset} summarizes the specifications of each dataset.  We report the average of accuracy, processing time, and maximum memory consumption for each dataset in Table \ref{tab:accuracy}, \ref{tab:time}, and \ref{tab:memory}, respectively. 

\subsubsection{Synthetic Datasets }
According to the concept drift scenarios of Zhang et al. \cite{bib:zhang}, we have 8 Rigorous Concept Drifting (RCD) and 8 Loose Concept Drifting (LCD) synthetic datasets. Bifet et al. \cite{bib:2009} specified Random RBF generator as the RCD data stream, and the rest of synthetic data stream generators as the LCD data stream. 

\textit{\textbf{Random RBF.}} It assigns a fixed number of random positioned centroids, with a random standard deviation value, class label, and weight. For generating new instances, we randomly select a center, considering weights, so that centroids with higher weights are more likely to be chosen. A random direction is chosen for displacement, using a Gaussian distribution, and drift is defined by moving the centroids, with constant speed. Attributes are all numerical values. Using this generator we prepared 8 different datasets, each containing 1 million instances, with 20 attributes, and 0 percent noise. Here are 3 important alternate factors we changed among these 8 datasets. We reflect these, respectively, in the naming of RBF datasets in Table \ref{tab:Dataset}.

\begin{table}
	\tbl{Summary of Dataset Characteristics\label{tab:Dataset}}
	{
		\begin{tabular}{lccccc} 
		\toprule 
		Dataset & \#Instance & \#Att & \#CL & \%N & Drift Spec. \\ 
		\hline 
		RBF-G-4-S  & \( 1\times 10^6 \) & 20 & 4  & 0 & Gr., Bp., DS=0.0001\\
		RBF-G-4-F  & \( 1\times 10^6 \) & 20 & 4  & 0 & Gr., Bp., DS=0.01\\
		RBF-G-10-S & \( 1\times 10^6 \) & 20 & 10 & 0 & Gr., Bp., DS=0.0001\\
		RBF-G-10-F & \( 1\times 10^6 \) & 20 & 10 & 0 & Gr., Bp., DS=0.01\\ 	
		RBF-A-4-S  & \( 1\times 10^6 \) & 20 & 4  & 0 & Abrupt, \#D=10\\
		RBF-A-4-F  & \( 1\times 10^6 \) & 20 & 4  & 0 & Abrupt, \#D=100\\
		RBF-A-10-S & \( 1\times 10^6 \) & 20 & 10 & 0 & Abrupt, \#D=10\\
		RBF-A-10-F & \( 1\times 10^6 \) & 20 & 10 & 0 & Abrupt, \#D=100\\
		\midrule
		SEA-S & \( 1\times 10^6 \) & 3 & 2 & 10 & Abrupt, \#D=3\\
		SEA-F & \( 2\times 10^6 \) & 3 & 2 & 10 & Abrupt, \#D=9\\
		\midrule
		HYP-S & \( 1\times 10^6 \) & 10 & 2 & 5 & Incrm., DS=0.001\\
		HYP-F & \( 1\times 10^6 \) & 10 & 2 & 5 & Incrm., DS=0.1\\
		\midrule
		TREE-S & \( 1\times 10^6 \) & 10 & 4 & 0 & Reoc., \#D=4\\
		TREE-F & \( 1\times 10^5 \) & 10 & 6 & 0 & Reoc., \#D=15\\
		\midrule
		LED-M & \( 1\times 10^6 \) & 24 & 10 & 10 & Mixed, \#D=3\\
		LED-ND & \( 1\times 10^7 \) & 24 & 10 & 20 & No drift\\
		\midrule
		CoverType & 581,012 & 54 & 7 & - & Unknown\\
		PokerHand & \( 1\times 10^7 \) & 10 & 10 & - & Unknown\\
		CovPokElec & 1,455,525 & 72 & 10 & - & Unknown\\
		Airlines & 539,383 & 7 & 2 & - & Unknown\\
		
		\bottomrule 
		\end{tabular}
	}
	\begin{tabnote}
		\Note {\#CL:} {No. of Class Labels,} {\%N:} {Percentage of Noise,} {DS:} {Drift Speed,} {\#D:} {No. of Drifts,} {Gr.:} {Gradual,} {Bp.:} {Blips.}
	\end{tabnote}
	
\end{table}

\begin{itemize}
	\item \textit{Concept Drift Type (Gradual: G and Abrupt: A).} The way the generator moves centroids make the data stream gradually changing. We add some outliers during generations of gradual changing datasets in order to have blips. We generate abruptly changing data streams using the sigmoid join operator (\(c=a \oplus^W_{t_0} b\); \(t_0\): point of change, \(W\): length of change) \cite{bib:2009}.
	\item \textit{Number of Classes (Four: 4 and Ten: 10).} The ability to generate an arbitrary number of classes is useful for evaluating an algorithm. We generate our datasets with either four or ten class labels.
	\item \textit{Drift Frequency (Slow: S and Fast: F).} For gradually changing datasets, we generate instances with 0.01 (fast) and 0.0001 (slow) concept change speed (defined as moving centroids in a random direction for a predefined distance of 0.01 or 0.001, within each 500 instances). For abruptly changing datasets, we switch to a new random stream generator that generates data stream with zero concept changing speed, 10 (slow) or 100 (fast), times evenly distributed over 1 million instances.
\end{itemize}

\textit{\textbf{SEA Concepts.}} It involves 3 numerical attributes varying between 0 and 10 \cite{bib:sea}. In our experiment, we use this generator in 2 different settings, both with 10 percent noise. First, 1 million instances, with drifts occurring every 250,000 examples (slow: SEA-S), and second, 2 million instances with drifts occurring every 200,000 examples (fast: SEA-F) are generated.

\textit{\textbf{Rotating Hyperplane.}} It assigns points in a multi-dimensional hyperplane and classifies them positively and negatively. Concept drift is defined by changing the orientation and position of the hyperplane \cite{bib:rothyper}. We set the hyperplane generator to create 2 datasets, each with 1 million instances described by 10 numerical features. We add 5 percent class noise to both of them. The modification weight of slowly changing dataset (HYP-S) is set to \( w_i= 0.001 \), and for the rapidly changing one (HYP-F) to \( w_i= 0.1 \).

\textit{\textbf{Random Tree.}} It produces nominal and numerical attributes using a randomly constructed tree. Drift is defined by abruptly changing the tree after a given number of examples \cite{bib:moa}. For both slow and fast tree datasets, we set the generator to have 5 nominal and 5 numerical attributes. The slowly changing dataset (TREE-S) consists of 1 million instances, with 4 evenly distributed reoccurring drifts. The rapidly changing dataset (TREE-F) contains 100,000 instances with 15 sudden drifts; it is the fastest changing dataset in our experiments.

\textit{\textbf{LED.}} It tries to predict the digit displayed on a seven-segment LED display. Each instance has 24 binary attributes and each has a possibility of being inverted, which is defined as noise. We have 2 LED datasets. The first dataset, LED-M, has 1 million instances with 2 gradually drifting concepts abruptly switching after 0.5 million instances, and 10 percent noise. The second, LED-ND, has 10 million instances without any drift and 20 percent noise, making it the noisiest and largest dataset \cite{bib:aue}. 

\subsubsection{Real-World Datasets}
The noise values, number of drifts, and drift speeds are unknown for these datasets. Access URL links are given in the footnote. 

\textit{\textbf{CoverType.}}\footnote{ Access link: http://archive.ics.uci.edu/ml/datasets/Covertype } It contains the forest cover type from the US Forest Service (USFS), comprised of 581,012 instances and 54 attributes.   

\textit{\textbf{PokerHand.}}\footnote{ Access link: http://archive.ics.uci.edu/ml/datasets/Poker+Hand } It consists of 1 million instances and 10 attributes. Each record is a hand of 5 playing cards---with 2 attributes as suit and rank.

\textit{\textbf{CovPokElec.}}\footnote{ Access link: http://www.openml.org/d/149 } It combines the normalized CoverType, normalized PokerHand, and Electricity datasets using the sigmoid join operator. The Electricity dataset comes from the Australian New South Wales Electricity Market. CovPokElec is obtained by merging all attributes, and assuming that each dataset corresponds to a different concept \cite{bib:2009}.

\textit{\textbf{Airlines.}}\footnote{ Access link: http://moa.cms.waikato.ac.nz/datasets/ } It consists of 539,383 examples described by 7 attributes. The task is to predict whether or not a given flight will be delayed, given the information of the scheduled departure.

\subsection{Experimental Framework: Detailed Design}
\textit{Implementation details. }In this paper, we use the Massive Online Analysis (MOA)\footnote{MOA webpage: \url{http://moa.cms.waikato.ac.nz/} } framework \cite{bib:moa}. MOA is an open-source software package to run data streaming experiments and, to the best of our knowledge, is the most popular framework for data stream mining. We use the JAva MAtrix (JAMA)\footnote{JAMA webpage: \url{http://math.nist.gov/javanumerics/jama/} } package, a basic linear algebra library, for matrix operations and to find least squares solutions in our implementation of GOOWE. We extended MOA for GOOWE implementation using the Java programming language. Some of the other ensemble algorithms, we used as baselines; they are implemented as part of the MOA framework. We used the MOA extensions library for DWM and NSE. In addition, our implementation of GOOWE, and some detailed information about experimental evaluation, such as standard deviations, and dataset generations are available on our GitHub webpage\footnote{GOOWE webpage: https://hamedrab.github.io/GOOWE/ }. 

\textit{Experimental Analysis.} We first study the impact of the proposed weighting system on vote aggregation and ensemble maintenance using two scenarios. In both of these scenarios, we use a fixed block-based ensemble, while different weighting systems are implemented in parallel to the original weighting system. In this way, we may study a single impact factor, and cancel all other impact factors. Through this analysis, GOOWE's weighting system is compared to most similar block-based ensembles, i.e. AUE2, AWE, and DWM, and some other baselines based on GOOWE's weighting system. 

\textit{Comparative Study.} For our comparative study, we evaluate GOOWE by comparing it with 8 well-known ensemble classifiers for non-stationary environments using the online block-based, bagging, and boosting methods as baselines. We select AWE, AUE2, DWM, and NSE ensemble methods from block-based approaches. In addition to these, we include OAUE, OzaBag, OzaBoost, and LevBag ensemble methods as popular online ensembles proven to have reasonable performance in evolving environments. 

\textit{Ensemble Size.} As discussed in Section 2, ensemble size has an important impact on performance of different algorithms. We suggest in \cite{bib:bonabcikm} to have the same number of component classifiers as class labels. For our experimental analyses, we use the same number of classifiers as the number of class labels for each data stream. However, in order to ease the comparisons of time and memory consumption values, and to follow the convention in the literature of using a fixed maximum number of classifiers, we fixed ensemble size for our comparative study. We set the maximum number of classifiers to 10. Studies based on a fixed number of classifiers are acceptable, since in such cases all ensemble methods can be equally disadvantaged \cite{bib:bonabcikm}.

\textit{Base Classifier. }We use the Hoeffding tree \cite{bib:ht} as the base classifier component for all examined ensemble methods. We use the Hoeffding tree enhanced with adaptive Naive Bayes leaf predictions, with a grace period \(n_{min} = 100\), split confidence \( \delta = 0.01 \), and tie-threshold \( \tau = 0.05 \) similar to experiments in \cite{bib:aue,bib:oaue,bib:ht}. 

\textit{Chunk and Instance Window Size.} In our experiments, according to the chunk size analysis of \cite{bib:awe} and similar to the experimental evaluations of \cite{bib:aue}, the chunk size for block-based ensembles (namely DWM, NSE, AWE, AUE2, and GOOWE) is set to 500 instances. OAUE and GOOWE use a sliding window of recent data instances. To ensure a fair comparison, similar to block-based ensembles, we set the instance window length to 500 instances. Although this length can be smaller for most of the ensembles, to perform an equivalent comparison, we choose this value based on the suggested minimum chunk length of AWE \cite{bib:awe}. The data chunk size and instance window size analysis is possible as a future work.  

\textit{Measurements. }By considering the main requirements of data stream environments \cite{bib:2009,bib:aue,bib:sea} in our experimental setup, we chose the interleaved Test-Then-Train procedure for measuring prediction accuracy values. For time and memory measurements, we use CentiSecond (CS) and MegaByte (MB), respectively. Our initial experiments showed that for synthetic datasets with the exact same settings of data stream generators, accuracy, time, and memory measurements showed variations. In order to have confident conclusions, for each synthetic data stream, we generate 10 time-seeded random datasets. For example, when we say that RBF-G-4-F dataset has 1 million of instances, we examine 10 such datasets (i.e. a total of 10 millions of instances) and report the mean value among these 10. 

\textit{Machine Specification.} The experiments were performed on a machine equipped with an Intel Xeon E3-1200 v3 @ 3.40 GHz processor and 32 GB of ECC RAM.

\section{Experimental Analyses: the Impact of Weighting and Model Management Strategies of GOOWE}
In this section, we mainly focus on answering the question: why should GOOWE work better in terms of prediction accuracy, or to put it in other words, when/where in the learning process does GOOWE get its advantage? To answer this question, we need to study the impact of GOOWE's weighting system on vote aggregation and ensemble maintenance in evolving environments as two major features of GOOWE. These two features differentiate GOOWE from other block-based ensembles, and we show the superiority of GOOWE compared to other ensembles based on these two key features.  

We designed two scenarios for studying the impact of the weighting system of GOOWE on vote aggregation and ensemble maintenance. Detailed information regarding each of these scenarios are given in the following. The main idea in both analyses is that by isolating the examined feature the impact can be studied. We choose a basic and comparably good ensemble method, and fix all settings for training and testing, except for the studied one (vote aggregation or ensemble maintenance). Here for our analyses, we exploit AUE2 implementation from the MOA framework as the base ensemble, since the weighting system of other block-based ensembles can be applied easily; it is also one of the leading ensembles. For the following scenarios of analyses, we created two versions as \textit{Base1} and \textit{Base2}. \textit{Base1} includes every detail of the AUE2 ensemble, except its vote aggregation. \textit{Base2} includes every detail of the AUE2 ensemble, except decisions on add/drop components. Further explanations are provided for each of these in the following. Using these analyses, we can verify GOOWE's weighting system superiority without benefiting from other specifications of each ensemble. 

\begin{table*}
	\tbl{Classification Accuracy in Percentage (\%) for Vote Aggregation Analysis on Data Streams with Concept Drift---\textit{Base1} ensemble method with different weighting systems used for aggregating votes\label{tab:analysis1}}
	{	
		\begin{tabular}{lcccccccc} 
			\toprule 
			Dataset & MV & DWM ($\beta=0.5$) & DWM ($\beta=0.2$) & AWE & AUE2 & GOOWE & GOOWE-Min & GOOWE-Max\\			 
			\hline 
			RBF-G-4-S  & 31.854 & 31.983 & 31.989 & 30.834 & 31.214 & \textbf{33.853} & 29.692 & 33.627\\
			RBF-G-4-F  & 91.746 & 85.888 & 85.666 & 90.868 & 91.668 & \textbf{91.626} & 72.478 & 87.066\\			
			RBF-G-10-S & 15.444 & 14.857 & 14.867 & 14.674 & 15.036 & \textbf{17.395} & 13.444 & 15.733\\			
			RBF-G-10-F & 80.794 & 80.956 & 80.929 & 80.939 & 80.864 & \textbf{84.062} & 77.054 & 78.378\\
			RBF-A-4-S  & 93.040 & 90.476 & 89.851 & 90.099 & 93.037 & \textbf{92.983} & 71.727 & 90.768 \\
			RBF-A-4-F  & 93.737 & 90.110 & 90.498 & 93.794 & 93.699 & \textbf{93.627} & 70.295 & 91.008 \\
			RBF-A-10-S & 90.460 & 90.017 & 90.011 & 90.980 & 90.675 & \textbf{93.869} & 80.094 & 85.469 \\
			RBF-A-10-F & 89.402 & 88.474 & 87.950 & 86.842 & 89.572 & \textbf{92.622} & 80.934 & 84.916 \\
			\midrule
			SEA-S & 85.636 & 86.927 & 86.921 & 85.289 & 85.847 & \textbf{84.510} & 82.364 & 83.670 \\
			SEA-F & 89.433 & 89.302 & 89.288 & 89.520 & 89.230 & \textbf{89.409} & 86.289 & 89.310 \\
			HYP-S & 83.140 & 87.461 & 87.462 & 86.811 & 84.587 & \textbf{82.429} & 80.351 & 82.134\\
			HYP-F & 90.742 & 90.955 & 90.953 & 90.981 & 90.382 & \textbf{91.189} & 88.434 & 91.007\\
			TREE-S & 94.632 & 94.452 & 94.452 & 94.750 & 94.599 & \textbf{94.796} & 58.737 & 94.470\\
			TREE-F & 82.280 & 81.993 & 81.963 & 81.445 & 82.199 & \textbf{82.560} & 54.021 & 82.255\\
			LED-M  & 73.649 & 73.266 & 73.256 & 72.646 & 73.645 & \textbf{73.599} & 69.046 & 73.575 \\
			\midrule
			CoverType  & 86.516 & 87.841 & 87.881 & 85.655 & 83.306 & \textbf{88.139 }& 75.597 & 87.609\\
			PokerHand  & 66.851 & 68.707 & 68.441 & 66.451 & 66.826 & \textbf{71.823} & 60.204 & 63.281\\
			CovPokElec & 74.845 & 75.911 & 75.831 & 75.584 & 74.909 & \textbf{79.519} & 68.341 & 71.845\\
			Airlines   & 62.136 & 62.663 & 62.689 & 62.041 & 62.293 & \textbf{62.368} & 60.957 & 62.116\\	
			\bottomrule 
		\end{tabular}
	}
\end{table*}

\subsection{Analysis of Vote Aggregation}
For evaluating the impact of the weighting strategy proposed for GOOWE on vote aggregation, as previously described, we use the AUE2 implementation from the MOA framework, except for its vote aggregation, as the base ensemble method, called \textit{Base1}. We implement GOOWE's weighting system for the \textit{Base1} ensemble classifier. As a result, the only variant to this new ensemble, compared to the original AUE2 version, is our weighting system for vote aggregation. In this way, we are able to study the impact of any weighting function in vote aggregation on the accuracy of predictions. 

In order to have different vote aggregation rules as our baselines, we also implement Majority Voting (MV), DWM with punishment constant values of $0.5$ and $0.2$, and also AWE's weighting systems for \textit{Base1} ensemble. In addition, we include the prediction accuracy of the component corresponding to the least/highest weight obtained from GOOWE weighting system (in Table \ref{tab:analysis1} illustrated as GOOWE-Min and GOOWE-Max). GOOWE-Min presents the worst-performing component and GOOWE-Max presents the best performing component, according to GOOWE's weights. We conduct our analysis using these as state-of-the-art baselines of weighting systems.    

\begin{figure*}
	\centering
	\includegraphics[width=0.79\linewidth, height=5.36cm]{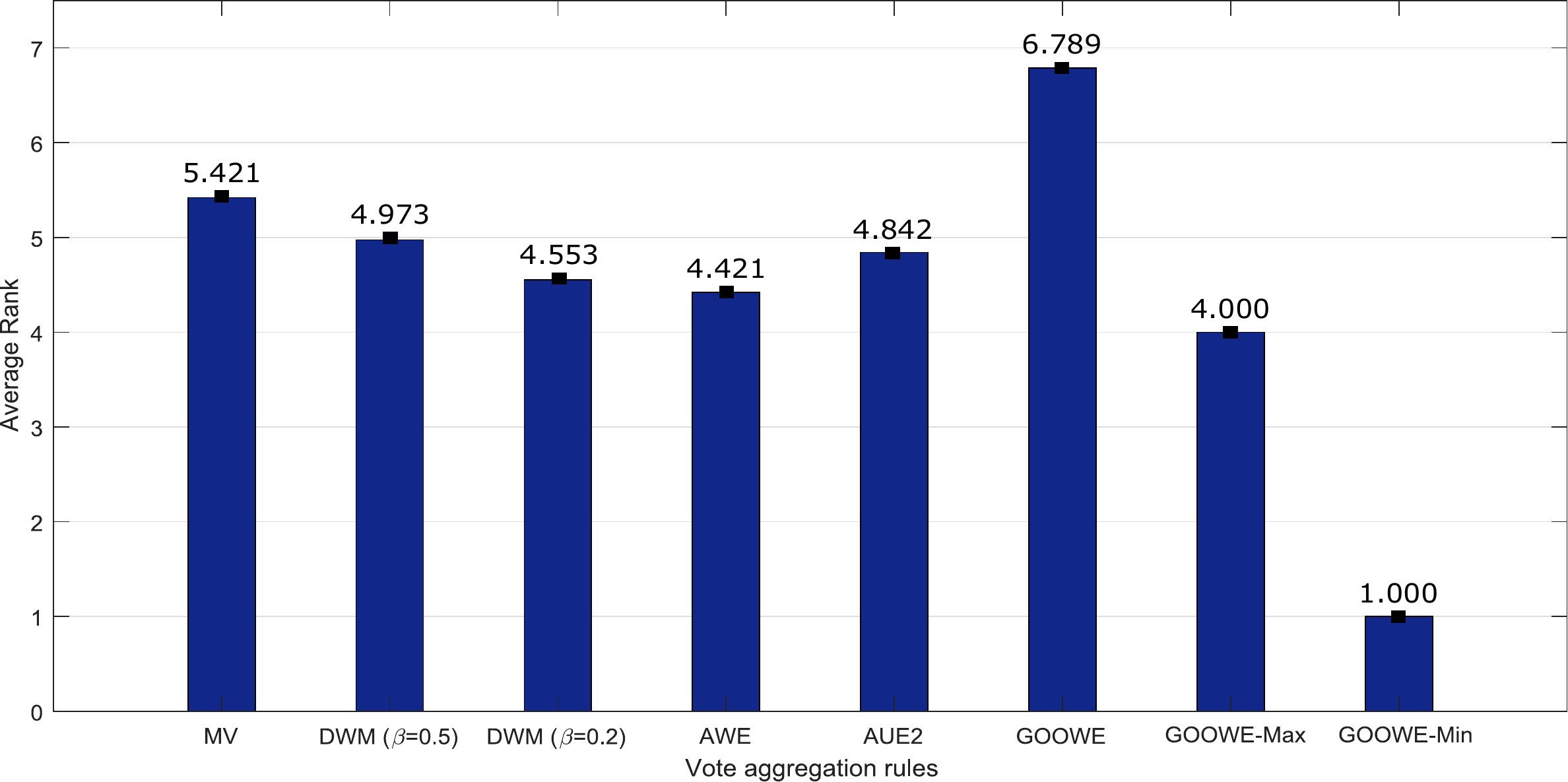}
	\caption{The Friedman statistical test average ranks for different vote aggregation rules. Higher average rank means better prediction accuracy. The minimum required difference of average ranks for considering a statistically significant difference is $1.197$.  } 
	\label{fig:anal1}
\end{figure*}

Table \ref{tab:analysis1} presents the accuracy values obtained from the mentioned vote aggregation rules. Note that in all of these scenarios, the data stream has concept drift. In order to compare these aggregation rules we conduct the non-parametric Friedman statistical test with pairwise comparisons. The null-hypothesis states that all aggregation rules are equal \cite{bib:statJournal,bib:statBook}. Since we have 8 vote aggregation rule and 19 datasets in our experiment, \(F_F\) is distributed according to the F distribution with \( 8-1=7 \) and \( (8-1) \times (19-1) = 126 \) degrees of freedom. We run the statistical test at the significance level of \( \alpha = 0.05\) and reject the null-hypothesis with a \textit{p-value} of $<0.00001$. 

The multiple comparisons average ranks are plotted in Figure \ref{fig:anal1}. The Critical Distances (CD) for \( F(8,152) = 14.802\) is $1.197$, meaning average ranks of aggregation rules need to have at least this amount of difference to be considered statistically significantly different, in a pairwise comparison. The weighting system of GOOWE is statistically significantly better compared to all other baseline aggregation rules, as shown in Figure \ref{fig:anal1}. While MV acts very well among remaining aggregation rules in evolving environments, we are not able to claim a statistically significant difference among them---excluding GOOWE-Max and GOOWE-Min.  

Based on our preliminary tests, GOOWE's weighting system shows its superiority in evolving environments. For this purpose, we tested our analysis scenario on RBF and LED data streams without any concept drift; there was no meaningful difference between MV and GOOWE's weighting systems. This is because when concept drift happens, GOOWE reacts much faster. The same can be concluded when we compare rapidly changing data streams with slowly changing ones, in Table \ref{tab:analysis1}. We will show this with more details through our comparative experiments in the next section.

\subsection{Analysis of Model Management Strategy}
For examining the superiority of our model management strategy, similar to the previous analysis, we use the implementation of AUE2 from the MOA framework, as \textit{Base2}. In this analysis, we use GOOWE and other baseline weights in the process of making decisions for add/drop components. We implement these baselines for the \textit{Base2} ensemble. Note that, for aggregating the votes of components in this analysis, we use majority voting to equally disadvantage all the ensembles. We use DWM with $\beta = 0.5$ and AWE as baselines of this analysis. DWM with $\beta = 0.2$ gave the exact same results as DWM with $\beta = 0.5$. We construct and maintain the \textit{Base2} ensemble using these weighting algorithms for each data stream. Table \ref{tab:analysis2} presents the resulting accuracies. 

\begin{table*}
	\tbl{Classification Accuracy in Percentage (\%) for the Model Management Analysis on Data Streams with Concept Drift---\textit{Base2} ensemble method with different weighting systems used for the decision of components add/drop\label{tab:analysis2}}
	{	
		\begin{tabular}{lcccc} 
			\toprule 
			Dataset & DWM & AWE & AUE2 & GOOWE\\			 
			\hline 
			RBF-G-4-S  & 34.077 & 30.846 & 31.854 & \textbf{32.110}\\
			RBF-G-4-F  & 89.424 & 89.99 & 91.746 & \textbf{92.176}\\			
			RBF-G-10-S & 17.404 & 14.104 & 15.444 & \textbf{15.241}\\			
			RBF-G-10-F & 89.992 & 80.161 & 80.794 & \textbf{90.995}\\			
			RBF-A-4-S  & 93.651 & 87.399 & 93.040 & \textbf{93.610}\\			
			RBF-A-4-F  & 94.310 & 87.265 & 93.737 & \textbf{94.389}\\
			RBF-A-10-S & 95.015 & 85.528 & 90.460 & \textbf{95.437}\\
			RBF-A-10-F & 95.395 & 86.230 & 89.402 & \textbf{95.259}\\
			\midrule
			SEA-S & 89.353 & 87.75 & 85.636 & \textbf{89.196}\\			
			SEA-F & 89.565 & 89.010 & 89.433 & \textbf{89.469}\\
			HYP-S & 86.779 & 83.726 & 83.140 & \textbf{83.009}\\
			HYP-F & 88.035 & 90.797 & 90.742 & \textbf{91.189}\\
			TREE-S & 94.813 & 94.632 & 94.632 & \textbf{84.529}\\
			TREE-F & 82.280 & 82.280 & 82.280 & \textbf{78.830}\\			
			LED-M  & 73.644 & 73.619 & 73.649 & \textbf{73.596}\\
			\midrule
			CoverType  & 88.204 & 87.344 & 86.516 & \textbf{88.004}\\		
			PokerHand  & 85.716 & 67.637 & 66.851 & \textbf{81.702}\\
			CovPokElec & 88.935 & 74.818 & 74.845 & \textbf{81.849}\\
			Airlines   & 64.570 & 63.084 & 62.136 & \textbf{62.146}\\	
			\bottomrule 
		\end{tabular}
	}
\end{table*}

In Table \ref{tab:analysis2} we observe a similar superiority of the GOOWE weighting system for rapidly changing data streams, compared to slowly changing data streams. The same scenario is valid here; GOOWE gets its advantage when more concept drifts happen, while reacting similarly in non-changing environments. 

Similarly to previous analysis, we conduct the non-parametric Friedman statistical test with pairwise comparisons. The null-hypothesis states that all model management strategies are equal. Since we have 4 algorithms and 19 datasets in our experiment, \(F_F\) is distributed according to the F distribution with \( 4-1=3 \) and \( (4-1) \times (19-1) = 54 \) degrees of freedom. We run the statistical test at the significance level of \( \alpha = 0.05\), and get \( F(3,54) = 8.3937\). We are able to reject the null-hypothesis with a \textit{p-value} of $0.0001$. Moreover, pairwise multiple comparisons indicate no statistically significant superiority for GOOWE, in ensemble maintenance, compared to DWM and its superiority compared to AUE2 and AWE. 

\textbf{\textit{Conclusion of the Experimental Analyses.}} Our first analysis shows the superiority of GOOWE vote aggregation in evolving environments. The second analysis shows GOOWE's conservative behavior in ensemble maintenance. We can conclude that GOOWE gets its advantage with vote aggregation, while reacting similarly as the best block-based ensembles for model management.

\begin{table*}
	\tbl{Average Classification Accuracy in Percentage (\%) 
		---for each synthetic dataset a one-way ANOVA using Scheffe multiple comparisons statistical test is conducted, and the top tier algorithms are underlined; for real-world datasets the most accurate algorithms are underlined\label{tab:accuracy}}
	{	
		\begin{tabular}{lccccccccc} 
			\toprule 
			Dataset & DWM & NSE & AWE & AUE2 & GOOWE & OAUE & OzaBag & LevBag & OzaBoost \\ 
			\hline 
			RBF-G-4-S  & 75.157 &	72.355 & 75.329 & \underline{91.174} & \underline{92.014} & \underline{91.817} & 87.084 & 85.779 & 88.353 \\
			RBF-G-4-F  & 74.102 &	72.041 & 73.837 & \underline{94.250} & \underline{94.590} &	\underline{93.322} & 87.213 &	85.947 & 87.995 \\
			RBF-G-10-S & 79.549 & 77.365 & 81.326 & 83.102 & \underline{92.298}	& 83.059 & 80.901 & 80.671 & 76.951 \\
			RBF-G-10-F & 79.669 & 78.455 & 80.875 & 83.055 & \underline{92.189} & 82.726 & 80.748 & 80.256 & 76.459 \\ 	
			RBF-A-4-S  & 76.628 & 73.308 & 78.046 & \underline{96.543} & \underline{96.901} & \underline{96.267} & \underline{95.618} & \underline{95.676} & \underline{97.367} \\
			RBF-A-4-F  & 75.452 & 72.519 & 77.591 & \underline{96.779} & \underline{97.019} & \underline{95.867} & \underline{95.461} & \underline{95.988} & \underline{96.258} \\
			RBF-A-10-S & 81.297 & 79.446 & 84.832 & 91.943 & \underline{96.477} & 85.771 & \underline{95.017} & \underline{94.901} & \underline{95.136} \\
			RBF-A-10-F & 82.338 & 80.471 & 85.657 & 92.592 & \underline{96.730} & 88.473 & \underline{95.504} & \underline{95.480} & \underline{95.923} \\
			\midrule
			SEA-S & 86.030 & 86.847 & 87.897 & \underline{89.718} & 89.031 & \underline{89.749} & 89.628 & 89.633 & 89.360 \\
			SEA-F & 86.084 & 86.849 & 87.923 & \underline{89.812} & 89.637 & \underline{89.831} & 89.739 & 89.742 & 89.551 \\
			\midrule
			HYP-S & 86.819 & 87.175 & \underline{90.483} & 88.486 & \underline{88.891} & \underline{89.044} & 83.467 & \underline{89.222} & 86.306 \\
			HYP-F & 90.734 & 88.714 & 90.994 & \underline{92.564} & \underline{92.567} & \underline{92.748} & 82.032 & 92.148 & 89.495 \\
			\midrule
			TREE-S & 32.921 & 24.638 & 33.926 & 35.222 & 35.932 & 36.286 & \underline{37.135} & \underline{37.124} & 24.639 \\
			TREE-F & 28.870 & 09.512 & 30.154 & 31.858 & \underline{33.827} & 32.024 & 32.217 & 32.217 & 9.518 \\
			\midrule
			LED-M  & 65.880 & 70.354 & \underline{74.002} & \underline{73.975} & \underline{73.989} & \underline{73.973} & \underline{73.984} & \underline{74.008} & 73.778 \\
			LED-ND & 43.336 & 46.849 & \underline{51.210} & \underline{51.196} & \underline{51.191} & \underline{51.208} & \underline{51.194} & \underline{51.212} & 51.034 \\
			\midrule
			CoverType  & 86.266 & 79.437 & 80.600 & 84.951 & 89.779 & 88.205 & 84.264 & 84.080 & \underline{90.570} \\
			PokerHand  & 47.591 & 49.550 & 49.470 & 50.217 & \underline{54.604} & 50.031 & 52.995 & 52.995 & 53.681 \\
			CovPokElec & 85.377 & 65.249 & 66.330 & 73.668 & \underline{88.718} & 86.390 & 82.265 & 77.647 & 87.154 \\
			Airlines   & 61.206 & 60.797 & 60.650 & 61.395 & 61.834 & \underline{62.516} & 61.404 & 62.164 & 61.015 \\
			
			\bottomrule 
		\end{tabular}
	}
\end{table*}

\section{Comparative Evaluation}
In this section we examine GOOWE as an ensemble algorithm, as described in Algorithm 1, and compare it with the 8 most state-of-the-art ensemble methods. We measure class label prediction accuracy (in percentage), maximum memory usage (in MegaByte), and total processing time of every one thousand instances (in CentiSecond) for each of the ensemble algorithms---average values for synthetic datasets and exact values for real-world datasets reported in Table \ref{tab:accuracy}, \ref{tab:time}, and \ref{tab:memory}, respectively. For each synthetic dataset, a one-way analysis of variance (ANOVA) using Scheffe multiple comparisons \cite{bib:scheffe} are conducted, and the best-performing algorithms are underlined. It is not possible to conduct the Scheffe statistical test for real-world datasets, since they only have a single value. For each of them, we underline the most accurate and least resource consuming algorithm. 

\begin{figure}
	\centering
	\includegraphics[width=0.95\linewidth]{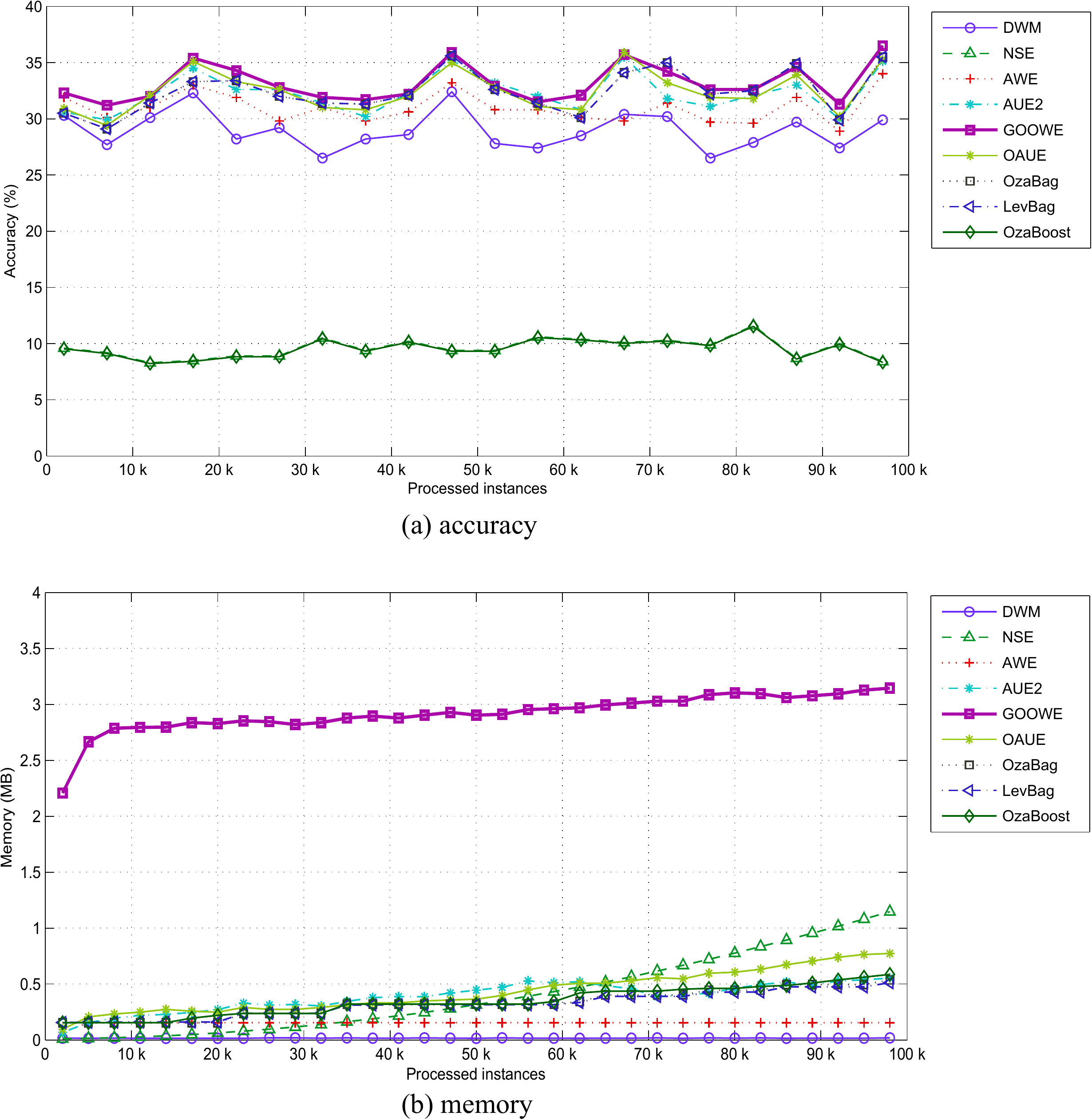}
	\caption{LCD example with reoccurring data stream: Classification accuracy and memory consumption for TREE-F dataset. } 
	\label{fig:tree}
\end{figure}

\begin{figure}
	\centering
	\includegraphics[width=0.95\linewidth]{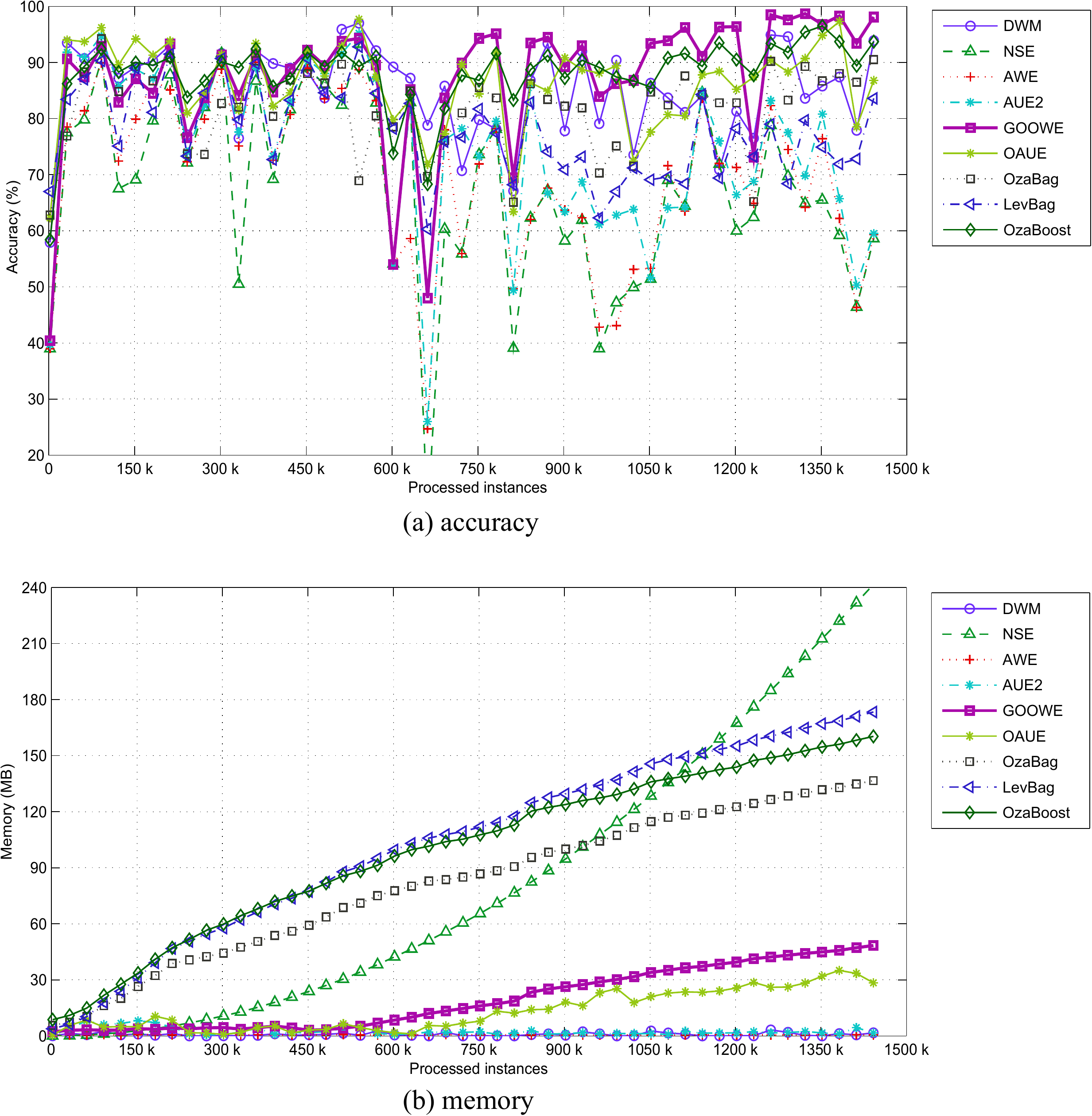}
	\caption{Real-world example data stream: Classification accuracy and memory consumption for CovPokElec dataset. } 
	\label{fig:CovPokElec}
\end{figure}

We draw scatter diagrams of the algorithms on the arrival of new chunks of data streams, as in \cite{bib:2009,bib:nse,bib:aue}. We provide one plot of accuracy and memory behavior for each category of RCD, LCD, and real-world datasets. For better understanding the behavior of ensembles in these situations, we present accuracy and memory plots for gradual changing RCD, and abrupt changing RCD datasets, separately. We provide these plots in Fig. \ref{fig:RBFG}, \ref{fig:RBFA}, \ref{fig:tree}, and \ref{fig:CovPokElec}---note that the plots are in different scales. 

\subsection{ RCD Data Streams with Gradual/Abrupt Drift Patterns} 
Table \ref{tab:accuracy} for Random RBF data streams (the first 8 rows) shows the superiority of GOOWE over other algorithms, in terms of accuracy. Its superiority is more significant for the gradually changing data streams, with respect to the abruptly changing data streams. Comparing the number of class labels suggests that GOOWE performs better for RCD datasets with 10 class-labels, rather than 4. The preliminary experiments show that this relationship changes with the number of component classifiers of the ensemble. For example, having 4 component classifiers can benefit more from a data stream with 4 class labels. 

\begin{table*}
	\tbl{Average Processing Time in CentiSecond (CS), for processing every one thousand instances---for each synthetic dataset a one-way ANOVA using Scheffe multiple comparisons statistical test is conducted, and the top tier algorithms are underlined; for real-world datasets the least time-consuming algorithms are underlined\label{tab:time}}
	{
		\begin{tabular}{lccccccccc} 
			\toprule 
			Dataset & DWM & NSE & AWE & AUE2 & GOOWE & OAUE & OzaBag & LevBag & OzaBoost \\ 
			\hline 
			RBF-G-4-S  & \underline{6.718}  & 439.071  & 18.458 & 20.563 & 14.887 & 17.482 & 8.299  & 17.825 & \underline{7.231} \\
			RBF-G-4-F  & \underline{6.628}  & 444.261  & 21.930 & 21.039 & 14.400 & 17.986 & 11.234 & 18.801 & 14.403 \\
			RBF-G-10-S & 31.854 & 1085.484 & 41.223 & 45.602 & 31.569 & 37.233 & \underline{17.146} & 38.402 & \underline{14.871} \\
			RBF-G-10-F & 30.501 & 1124.648 & 49.549 & 47.105 & 31.294 & 37.599 & \underline{23.036} & 38.789 & 28.333 \\ 	
			RBF-A-4-S  & \underline{6.598}  & 443.676  & 21.900 & 17.218 & 13.784 & 15.367 & 9.901  & 16.425 & 12.124 \\
			RBF-A-4-F  & \underline{6.621}  & 445.520  & 21.913 & 17.037 & 13.806 & 15.376 & 10.007 & 16.381 & 11.921 \\
			RBF-A-10-S & 30.027 & 1125.125 & 49.025 & 44.556 & 31.202 & 36.867 & \underline{21.545} & 36.214 & 27.433 \\
			RBF-A-10-F & 29.693 & 1119.683 & 48.678 & 43.537 & 30.718 & 36.231 & \underline{20.482} & 35.397 & 26.59 \\
			\midrule
			SEA-S & \underline{0.615} & 43.714 & 2.752 & 2.772 & 2.428 & 2.760 & 1.509 & 2.840 & 1.808 \\
			SEA-F & \underline{0.616} & 94.267 & 2.740 & 3.065 & 2.452 & 2.958 & 1.684 & 3.102 & 1.963 \\
			\midrule
			HYP-S & \underline{1.504} & 115.595 & 7.390 & 6.617 & 6.321 & 6.547 & 3.955 & 5.103 & 4.241 \\
			HYP-F & \underline{1.443} & 79.532  & 7.534 & 5.680 & 5.707 & 5.877 & 3.580 & 4.290 & 3.873 \\
			\midrule
			TREE-S & \underline{6.337} & 85.346 & 11.468 & 9.479  & 9.333  & 9.528  & 7.217 & 8.777 & 7.009 \\
			TREE-F & \underline{8.193} & 14.276 & 13.639 & 11.026 & 10.518 & 11.036 & \underline{7.641} & 9.478 & \underline{7.295} \\
			\midrule
			LED-M  & 35.260 & 1288.899 & 57.355 & 54.824 & 45.966 & 40.135 & \underline{13.874} & 25.615 & \underline{14.764} \\
			LED-ND & 38.603 & 1280.011 & 58.002 & 54.971 & 45.615 & 42.924 & \underline{16.068} & 23.699 & \underline{17.102} \\
			\midrule
			CoverType  & \underline{11.526} & 335.569 & 28.269 & 24.409 & 21.858 & 21.952 & 11.721 & 18.868 & 14.359 \\
			PokerHand  & 8.578  & 68.910  & \underline{5.184}  & 7.123  & 6.842  & 8.816  & 5.634  & 8.012  & 6.148 \\
			CovPokElec & 15.483 & 546.412 & 25.803 & 24.005 & 24.076 & 24.832 & \underline{13.786} & 18.814 & 17.804 \\
			Airlines   & \underline{0.982}  & 11.304  & 2.692  & 2.710  & 3.338  & 3.146  & 2.086  & 2.673  & 2.676 \\
			
			\bottomrule 
		\end{tabular}
	}
\end{table*}

\begin{table*}
	\tbl{Maximum Memory Usage in MegaByte (MB)
		---for each synthetic dataset a one-way ANOVA using Scheffe multiple comparisons statistical test is conducted, and the top tier algorithms are underlined; for real-world datasets the least memory-consuming algorithms are underlined\label{tab:memory}}
	{	
		\begin{tabular}{lccccccccc} 
			\toprule 
			Dataset & DWM & NSE & AWE & AUE2 & GOOWE & OAUE & OzaBag & LevBag & OzaBoost \\ 
			\hline 
			RBF-G-4-S  & \underline{0.261} & 116.193 & 0.320 & 0.483 &  1.632 & 0.467 & 13.379 & 15.221 & 13.232 \\
			RBF-G-4-F  & \underline{0.294} & 116.193 & \underline{0.319} & 2.885 &  9.750 & 2.711 & 35.258 & 39.568 & 24.517 \\
			RBF-G-10-S & \underline{0.333} & 116.201 & \underline{0.394} & 3.470 & 11.066 & 3.358 & 18.063 & 20.293 & 8.085 \\
			RBF-G-10-F & \underline{0.691} & 116.200 & \underline{0.394} & 5.451 & 11.837 & 5.408 & 16.877 & 17.382 & 13.343 \\ 	
			RBF-A-4-S  & \underline{0.931} & 116.192 & \underline{0.319} & 4.535 & 15.657 & 4.687 & 27.618 & 42.854 & 30.531 \\
			RBF-A-4-F  & \underline{0.553} & 116.192 & \underline{0.319} & 5.004 & 12.107 & 5.026 & 26.646 & 40.845 & 28.919 \\
			RBF-A-10-S & \underline{0.569} & 116.200 & \underline{0.395} & 4.453 & 11.320 & 4.194 & 19.184 & 35.068 & 16.292 \\
			RBF-A-10-F & \underline{0.842} & 116.200 & \underline{0.394} & 4.682 & 13.784 & 5.334 & 17.168 & 34.660 & 15.368 \\
			\midrule
			SEA-S & \underline{0.181} & 116.109 & \underline{0.229} & 18.974 & 3.975 & 21.146 & 7.046 & 9.009 & 6.861 \\
			SEA-F & \underline{0.204} & 464.351 & \underline{0.230} & 36.584 & 6.332 & 42.023 & 13.826 & 15.858 & 13.629 \\
			\midrule
			HYP-S & \underline{0.384} & 116.145 & \underline{0.614} & 2.376 & 4.849 & 2.636 & 22.918 & 25.218 & 21.19 \\
			HYP-F & \underline{0.505} & 116.147 & \underline{0.643} & 1.063 & 2.999 & 1.863 & 20.912 & 22.882 & 21.097 \\
			\midrule
			TREE-S & \underline{0.172} & 116.137 & \underline{0.210} & 1.426 & 5.931 & 7.343 & 6.288 & 9.265 & 6.343 \\
			TREE-F & \underline{0.181} & 1.251 & \underline{0.224} & 0.681 & 3.118 & 0.762 & 0.538 & 0.564 & 0.606 \\
			\midrule
			LED-M  & \underline{0.788} & 116.222 & \underline{0.453} & \underline{0.453} & \underline{1.052} & \underline{0.415} & 13.707 & 18.715 & 13.927 \\
			LED-ND & \underline{0.581} & 116.221 & \underline{0.453} & \underline{0.453} & \underline{2.871} & \underline{0.454} & 12.891 & 20.912 & 12.863 \\
			\midrule
			CoverType  & 1.252 & 39.376 & \underline{0.437} & 0.736 & 1.128 & 1.079 & 55.631 & 71.475 & 62.185 \\
			PokerHand  & \underline{0.186} & 116.146 & 0.224 & 0.254 & 0.924 & 0.229 & 3.460 & 4.390 & 3.524 \\
			CovPokElec & 4.534 & 246.144 & \underline{0.576} & 11.543 & 54.435 & 36.529 & 132.399 & 162.592 & 153.337 \\
			Airlines   & \underline{0.131} & 33.825 & 0.157 & 0.300 & 2.507 & 0.535 & 7.592 & 10.322 & 7.773 \\
			
			\bottomrule 
		\end{tabular}
	}
\end{table*}

\begin{figure}
	\centering
	\includegraphics[width=0.95\linewidth]{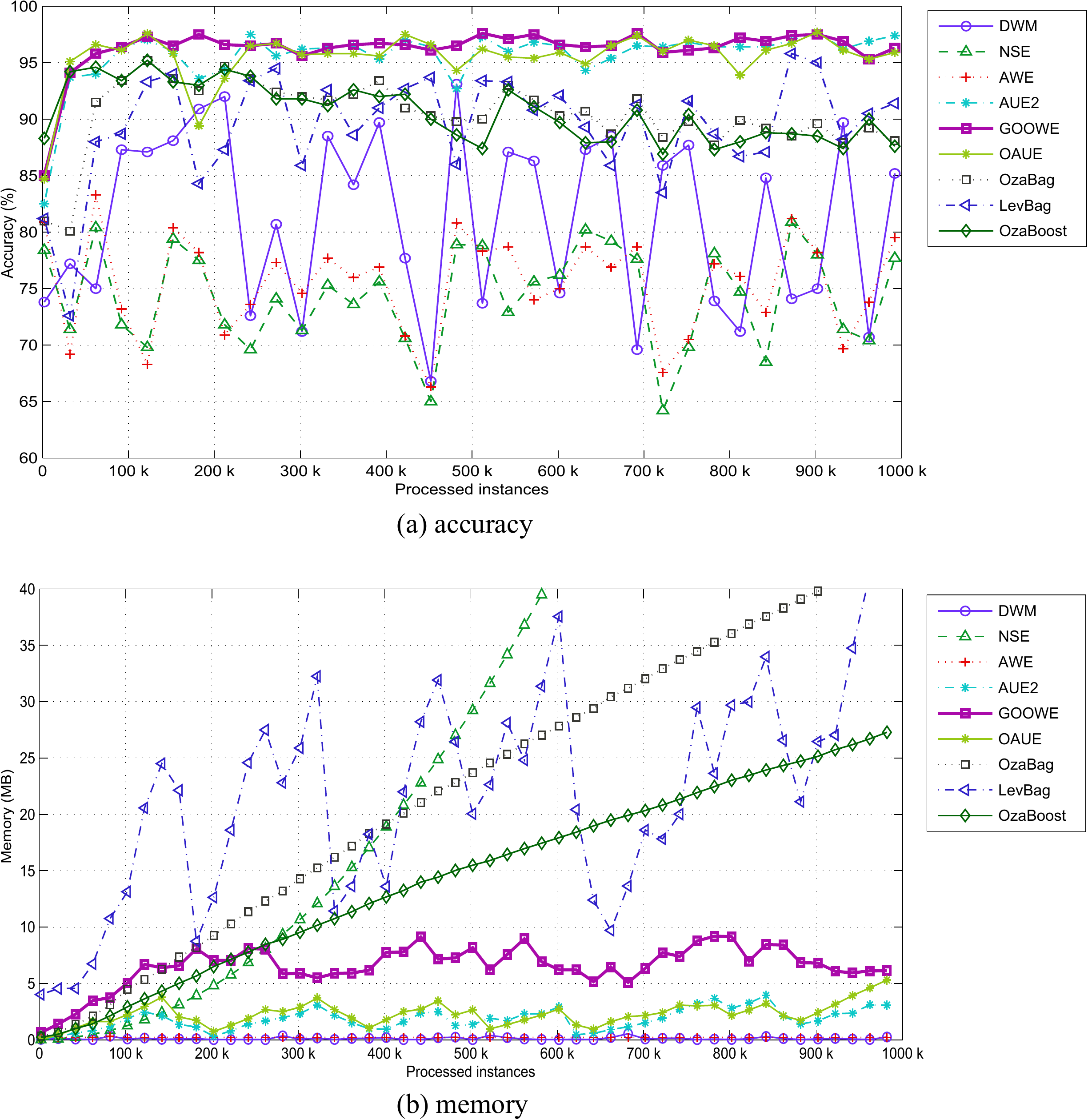}
	\caption{RCD example with a gradually changing data stream: Classification accuracy and memory consumption for RBF-G-4-F dataset. } 
	\label{fig:RBFG}
\end{figure}

\begin{figure}
	\centering
	\includegraphics[width=0.95\linewidth]{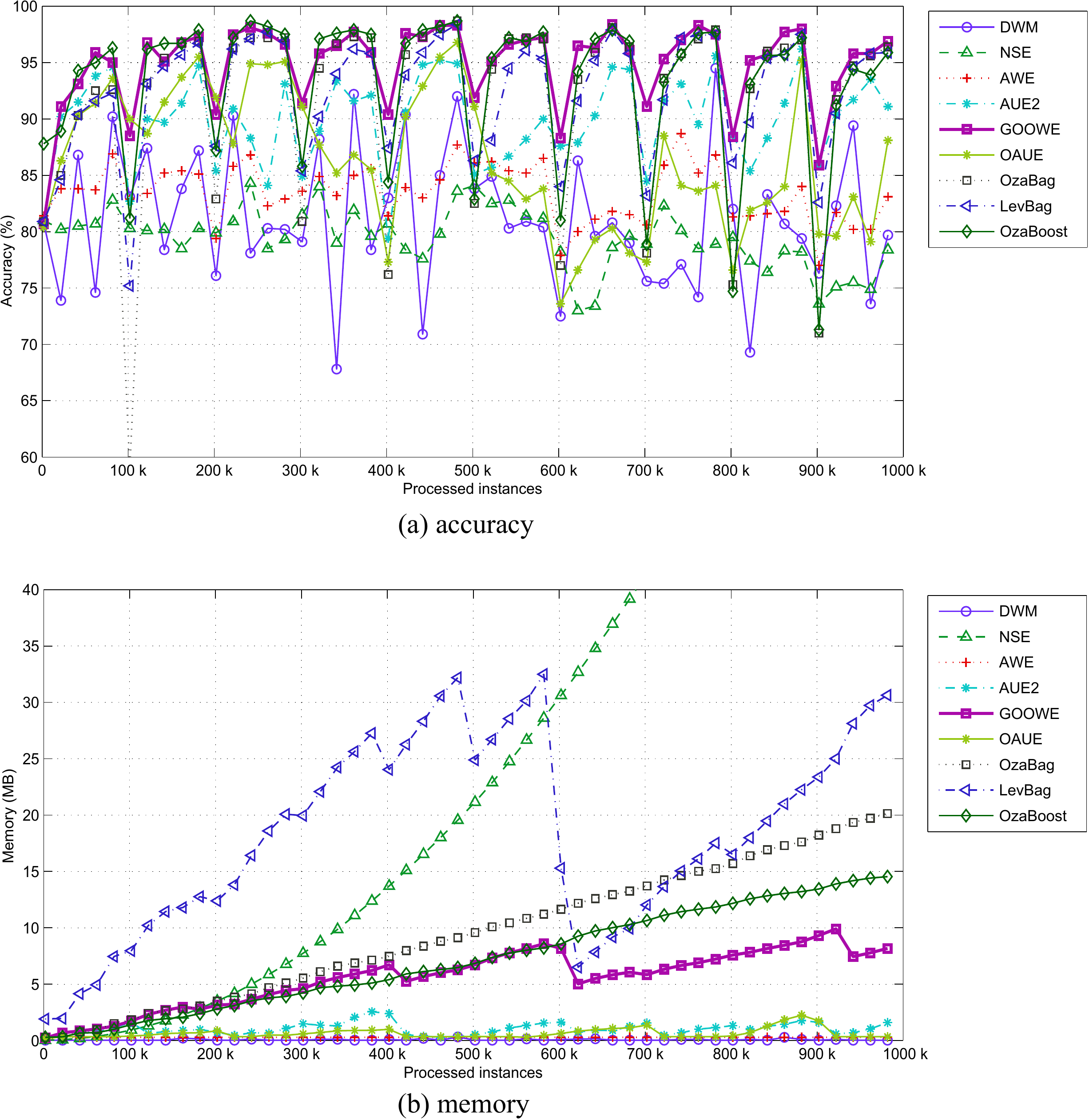}
	\caption{RCD example with abruptly changing data stream: Classification accuracy and memory consumption for RBF-A-10-S dataset. } 
	\label{fig:RBFA}
\end{figure}

As shown in Table \ref{tab:accuracy}, in most of the cases, GOOWE has higher average accuracy for the fast-changing datasets, compared to the slow changing ones. We can intuitively understand the reason in accuracy plots of Fig. \ref{fig:RBFG}-(a) and \ref{fig:RBFA}-(a). They present behaviors of different ensemble methods with the arrival of new data chunks of gradually/abruptly changing RBF data streams. The place of abrupt drifts is clear in the classification accuracy plots, consistent with what we know from the generation step of these synthetic datasets. In most abruptly changing points, it is obvious that GOOWE has significantly faster adaptive reactions than the others. While OzaBoost, LevBag and OzaBag perform similarly to GOOWE in stationary phases of the data stream, they react slowly in changing phases. As a result, when more changes exist in stream data, GOOWE provides better performance. DWM, NSE and AWE are among the poorly performing algorithms. 


Table \ref{tab:time} and \ref{tab:memory} for the RBF datasets (the first 8 rows) show the conservative resource consumption of GOOWE, in terms of time and memory. We present memory usage behavior for the algorithms in RBF-G-4-F and RBF-A-10-S datasets in Fig. \ref{fig:RBFG}-(b) and \ref{fig:RBFA}-(b). They show that most ensemble methods drop one of the most memory-hungry component classifiers with drift occurrence. Among these algorithms, the memory consumption of GOOWE is less than those of NSE, LevBag, OzaBag, and OzaBoost. Although it uses more memory than DWM, AWE, AUE2, and OAUE, it does not grow exponentially. As Brzezinski explained in \cite{bib:aue}, no pruning was used to limit the number of components for NSE, and it requires much more time and memory than the other algorithms. As a result, memory usage of NSE does not react to concept drifts and grows exponentially with the arrival of new instances.

\subsection{ LCD Data Streams with Miscellaneous Drift Patterns } 
Table \ref{tab:accuracy} for the LCD generators (the second 8 rows) shows that GOOWE is among the top-tier algorithms in the Rotating Hyperplane, TREE-F, and LED datasets, in terms of accuracy. Similarly, to the gradually changing RBF datasets, the Rotating Hyperplane dataset has incremental drifts. TREE-F is the smallest and fastest changing dataset with reoccurring drift patterns. These characteristics show the superiority of GOOWE. Generally, for the LCD datasets, we can say that GOOWE acts better for the fast-changing datasets, compared to the slow ones. For the LED datasets there is no significant difference among various algorithms, and most of them reacting similarly, since there are no clear concept drifts. For the SEA datasets, although GOOWE is not among the top tier algorithms, the differences among accuracy values are small. Moreover, Table \ref{tab:time} and \ref{tab:memory} for LCD data streams (second 8 rows) show comparable resource consumption of GOOWE, in terms of time and memory, similar to the RCD data streams.  

We can see the behavior of the algorithms for the TREE-F dataset, in terms of accuracy (Fig. \ref{fig:tree}-(a)) and memory usage (Fig. \ref{fig:tree}-(b)). Similarly to other accuracy plots, GOOWE reacts robustly to concept drifts. The memory usage plot suggests that GOOWE is the worst memory consumer algorithm. However, its memory growth rate is slow and its maximum memory usage is under 4 MB. In other words, it uses a limited amount of memory. In contrast to other plots, in Fig. \ref{fig:tree}-(b), NSE shows a small consumption of memory. In general, for NSE on small datasets, when only a few components are created, memory usage is reasonable.

\subsection{ Real-World Data Streams with Unknown Drift Patterns  }
Table \ref{tab:accuracy}, for real-world datasets (the last 4 rows), shows the superiority of GOOWE over other algorithms in PokerHand and CovPokElec datasets, in terms of accuracy. For CoverType and Airlines datasets, although GOOWE is not the best performing algorithm, still the difference with the best performing algorithms are less than 1 percent. In addition, Table \ref{tab:time} and \ref{tab:memory} for real-world datasets (the last 4 rows) show reasonable resource consumption of GOOWE, in terms of time and memory.    

Fig. \ref{fig:CovPokElec} shows classification accuracy and memory usage behaviors for the CovPokElec real-world dataset with the arrival of new data chunks. The accuracy plot (Fig. \ref{fig:CovPokElec}-(a)) shows that GOOWE, OzaBoost, OAUE and DWM are among the best performing ensemble methods. By tracking the behavior of different algorithms, in different situations, it is obvious that there are diverse types of concept drift in the CovPokElec dataset. For example, comparing the behavior of the algorithms around 350k and 400k demonstrates that, although all of the algorithms prove the existence of concept drift, for the first evolving point OzaBoost reacts best, in contrast to the second point, where DWM shows the best reaction. By looking at 750k or 1050k points, we can say that, in some situations, different algorithms are not synchronous; while some of them (DWM, OzaBag, LevBag) show a decrease in performance, the others (NSE, AWE, AUE2, GOOWE, OAUE, OzaBoost) show an increase. Considering the first 500k of instances, belonging to the normalized CoverType dataset, DWM and OAUE outperform the others and react faster to unknown drift types. For the second 500k of instances, belonging to the normalized PokerHand dataset, we see more robust behavior from GOOWE and OAUE. However, for the last 500k of instances, belonging to the Electricity dataset, except one evolving point around 1250k, the best performing algorithm is GOOWE. 

The memory plot (Fig. \ref{fig:CovPokElec}-(b)) suggests that while AWE, DWM and AUE2 are the least memory consumers, OAUE and GOOWE algorithms are far better than NSE, OzaBag, LevBag and OzaBoost which grow exponentially.

\begin{figure*}[t]
	\centering
	\includegraphics[width=0.8\linewidth]{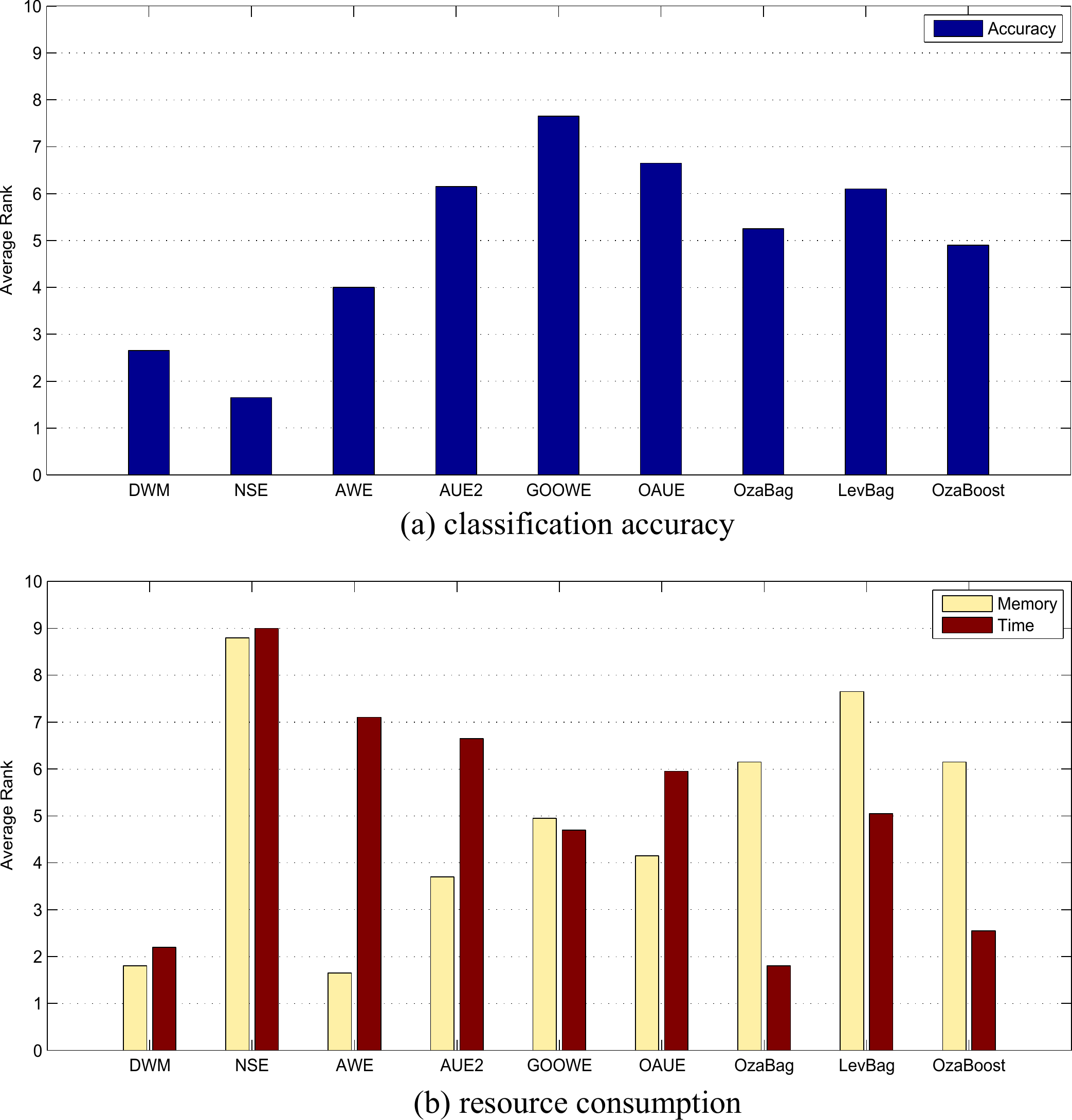}
	\caption{The Friedman statistical test average rank plots; for classification accuracy plot (a) higher average rank means better prediction, and for resource consumption plot (b) lower average ranks mean better performance. } 
	\label{fig:rankPlots}
\end{figure*}

\section{Statistical Analysis and Further Discussion}
In order to assure the significant difference of average values for classification accuracies, processing time, and memory usage, we carried out statistical tests. First, a one-way analysis of variance (ANOVA) test using Scheffe multiple comparisons \cite{bib:scheffe} were conducted on the results of different algorithms for each dataset. The null-hypothesis for each dataset when considered individually is: There is no significant difference between the algorithms.  

We conducted the Scheffe test at the significance level of \( \alpha = 0.05\). The most accurate, least consumer of processing time, and least consumer of memory algorithms are underlined for each row of Table \ref{tab:accuracy}, \ref{tab:time}, and \ref{tab:memory}. We underlined the top tier group of the Scheffe's comparison results for each synthetic dataset \cite{bib:scheffe}. As we mentioned earlier, it is not possible to conduct the Scheffe statistical test for real-world datasets, since they only have a single value. For each of them, we underline the most accurate and least resource consuming algorithm. As shown in Table \ref{tab:accuracy}, for 15 out of 20 datasets, GOOWE is consistently among the most accurate algorithms. OAUE is placed in the second rank of the most accurate algorithms, with 11 out of 20 datasets. For the cases of time and memory usage, Table \ref{tab:time} and \ref{tab:memory}, we can see that GOOWE is among the conservative consumers of resources. Despite this fact, when we compare resource usage with the worst ones, we can see that the costs are much less, and are affordable. In addition, comparing the resource usage of OAUE and LevBag with GOOWE shows that our algorithm is in the same range of memory and time consumption.

To extend the analysis, we carried out the non-parametric Friedman statistical test for comparing multiple classifiers over multiple datasets \cite{bib:statJournal,bib:statBook}. The null-hypothesis for this test states that all the algorithms are equivalent on all datasets when considered together. Since we have 9 algorithms and 20 datasets in our experiment, \(F_F\) is distributed according to the F distribution with \( 9-1=8 \) and \( (9-1) \times (20-1) = 152 \) degrees of freedom. We run the statistical tests at the significance level of \( \alpha = 0.05\); and the Critical Distances (CD) for \( F(8,152)\), and average ranks of algorithms are given in Table \ref{table:stat}. If the Friedman test results in a p-value less than  \( \alpha\), the null-hypothesis is rejected and we can conclude that at least 2 of the algorithms are significantly different from each other. The tests for accuracy, memory, and time results in p-values less than $0.00001$, and the null-hypotheses are rejected for all cases. We plot these average ranks in Fig. \ref{fig:rankPlots}. Note that for classification accuracy, Fig. \ref{fig:rankPlots}-(a), higher average rank means better prediction; and for resource consumption, Fig. \ref{fig:rankPlots}-(b), lower ranks mean better performance. 

Table \ref{table:stat} shows that, according to the Friedman test, GOOWE outperforms DWM, NSE, AWE, AUE2, OzaBag, LevBag, and OzaBoost, but not OAUE. The CD value is 1.238, and their rank difference is less than this value (\(7.650-6.650 = 1.000\)). Since the difference of average ranks between GOOWE and OAUE is close to the CD, we performed the Wilcoxon signed-rank test to further analyze this pair of algorithms \cite{bib:statJournal}. It ranks the absolute values of the differences between paired samples, and calculates a statistic on the number of negative and positive differences. For our case, the positive differences are 13, and the negative differences are 7. The two-tailed probability value, \( P = 0.014 \), is less than \( \alpha = 0.05\); Therefore, it can be accepted that GOOWE is significantly better than OAUE, in terms of accuracy. 

\begin{table*}[t]
	\tbl{Summary of the Friedman Statistical Test for Accuracy, Memory and Time; The underlined values are GOOWE and its rivals that are in the same range of rank with no significant difference \label{table:stat}}
	{	
		\begin{tabular}{lcccccccccccc}
			\toprule
			& \multicolumn{2}{c}{Test Results} & \multicolumn{9}{c}{Average Algorithm Ranks} \\
			\cmidrule(lr){2-3}\cmidrule(lr){4-12}
			Test         & \(F_F\) & CD & DWM & NSE & AWE & AUE2 & GOOWE & OAUE & OzaBag & LevBag & OzaBoost \\
			\midrule
			Accuracy       &  19.153  &  1.238  & 2.650 & 1.650 & 4.000 & 6.150 & \underline{7.650} & \underline{6.650} & 5.250 & 6.100 & 4.900 \\
			Memory         & 76.198  &  0.784  & 1.800 & 8.800 & 1.650 & 3.700 & \underline{4.950} & 4.150 & 6.150 & 7.650 & 6.150  \\
			Time           & 77.692  &  0.778  & 2.200 & 9.000 & 7.100 & 6.650 & \underline{4.700} & 5.950 & 1.800 & \underline{5.050} & 2.550  \\
			\bottomrule	
		\end{tabular}
	}
\end{table*}

Similar to the accuracy test, we performed time and memory statistical comparisons using the Friedman test, summarized in Table \ref{table:stat}. For the memory test, we reject the null-hypothesis. The average ranks show that GOOWE uses more memory compared to DWM, AWE, AUE2, and OAUE. On the other hand, it uses less memory compared to NSE, OzaBag, LevBag, and OzaBoost. The CD value shows that, GOOWE is in the middle of the baselines as a singleton, with a significant difference from the upper and lower range algorithms. For the processing time test, we again reject the null-hypothesis; the average ranks show that GOOWE is faster than NSE, AWE, AUE2, and OAUE. It is significantly slower than DWM, OzaBag,and OzaBoost, somehow equivalent to LevBag.

In summary, we can say that there is a trade-off between prediction accuracy and resource consumption. In this trade-off, GOOWE can predict statistically significantly better compared to the most accurate algorithms. Furthermore, it uses affordable resources compared to the most memory and time-efficient ensembles.

\section{Conclusions and Future Work}
In this paper, we provide a geometrically optimum and online-weighted ensemble classifier, called GOOWE, for non-stationary environments. The main contribution of the proposed algorithm is providing a spatial modeling for using the linear least squares (LSQ) solution for dynamically optimizing the weights of components of an ensemble classifier for evolving environments. Our algorithm, different from the use of LSQ in batch mode ensembles, has dynamically changing optimum weight assignment to component classifiers and continuous training and testing. We use data chunks for training and a sliding instance window containing the latest available data for testing; such an approach provides more robust behavior as shown in our experiments. We use the Euclidean norm as the measure of closeness in LSQ. The LSQ proved to react well in noisy situations \cite{bib:lsqbook}, as it did in our algorithm for data streams with concept drifts. 

First, we conduct an analysis for examining two major differentiating elements of GOOWE, the component weighting strategy and ensemble model management strategy. Our optimum and online weighting system shows its effectiveness for both vote aggregation and ensemble maintenance in evolving environments. Second, we experimentally compare GOOWE with 8 state-of-the-art ensemble classifiers, as our baselines, on 20 datasets as data streams with tens of millions of instances, where 16 of them are synthetic and 4 of them are real-world datasets. For the synthetic streams, we use two categories of data stream generators: Rigorous Concept Drift (RCD) and Loose Concept Drift (LCD), each with 8 datasets. We include all possible patterns of change (sudden/abrupt, incremental, gradual, and reoccurring as concept drifts, including blips and noise) in our datasets. The statistical tests prove the superiority and robustness of GOOWE in reacting to different types of concept drift, in terms of accuracy. Furthermore, we show that it requires conservative resource consumption, in terms of memory and processing time. 

In future work, the impact of the length of data chunk size on the performance in the presence of different concept drifts and its dynamic determination are possible studies.  Effects of instance window size on performance can also be analyzed. In addition, GOOWE can be used in various sub-problem domains, such as semi-supervised and multi-label classification. It can be applied to unbalanced datasets and streams with concept-evolution.    

\begin{acks}
We would like to thank Manouchehr Takrimi from Bilkent University, Jon M. Patton from Miami University of OH, and Alper Can for their valuable comments and pointers on  this paper.
\end{acks}

\bibliographystyle{ACM-Reference-Format-Journals}
\bibliography{acmsmall-sample-bibfile}


\begin{thebibliography}{00}


\ifx \showCODEN    \undefined \def \showCODEN     #1{\unskip}     \fi
\ifx \showDOI      \undefined \def \showDOI       #1{{\tt DOI:}\penalty0{#1}\ }
  \fi
\ifx \showISBNx    \undefined \def \showISBNx     #1{\unskip}     \fi
\ifx \showISBNxiii \undefined \def \showISBNxiii  #1{\unskip}     \fi
\ifx \showISSN     \undefined \def \showISSN      #1{\unskip}     \fi
\ifx \showLCCN     \undefined \def \showLCCN      #1{\unskip}     \fi
\ifx \shownote     \undefined \def \shownote      #1{#1}          \fi
\ifx \showarticletitle \undefined \def \showarticletitle #1{#1}   \fi
\ifx \showURL      \undefined \def \showURL       #1{#1}          \fi

\bibitem[\protect\citeauthoryear{Bifet and Gavald{\`{a}}}{Bifet and
  Gavald{\`{a}}}{2007}]%
        {bib:adwin}
{Albert Bifet} {and} {Ricard Gavald{\`{a}}}. 2007.
\newblock \showarticletitle{Learning from time-changing data with adaptive
  windowing}. In {\em Proceedings of the Seventh {SIAM} International
  Conference on Data Mining, April 26-28, 2007, Minneapolis, Minnesota, {USA}}.
  443--448.
\newblock


\bibitem[\protect\citeauthoryear{Bifet, Holmes, Kirkby, and Pfahringer}{Bifet
  et~al\mbox{.}}{2010b}]%
        {bib:moa}
{Albert Bifet}, {Geoff Holmes}, {Richard Kirkby}, {and} {Bernhard Pfahringer}.
  2010b.
\newblock \showarticletitle{{MOA:} Massive Online Analysis}.
\newblock {\em Journal of Machine Learning Research\/}  {11} (2010),
  1601--1604.
\newblock


\bibitem[\protect\citeauthoryear{Bifet, Holmes, and Pfahringer}{Bifet
  et~al\mbox{.}}{2010a}]%
        {bib:lev}
{Albert Bifet}, {Geoffrey Holmes}, {and} {Bernhard Pfahringer}. 2010a.
\newblock \showarticletitle{Leveraging bagging for evolving data streams}. In
  {\em Machine Learning and Knowledge Discovery in Databases, European
  Conference, {ECML} {PKDD} 2010, Barcelona, Spain, September 20-24, 2010,
  Proceedings, Part {I}}. 135--150.
\newblock


\bibitem[\protect\citeauthoryear{Bifet, Holmes, Pfahringer, Kirkby, and
  Gavald{\`{a}}}{Bifet et~al\mbox{.}}{2009}]%
        {bib:2009}
{Albert Bifet}, {Geoffrey Holmes}, {Bernhard Pfahringer}, {Richard Kirkby},
  {and} {Ricard Gavald{\`{a}}}. 2009.
\newblock \showarticletitle{New ensemble methods for evolving data streams}. In
  {\em Proceedings of the 15th {ACM} {SIGKDD} International Conference on
  Knowledge Discovery and Data Mining, Paris, France, June 28 - July 1, 2009}.
  139--148.
\newblock


\bibitem[\protect\citeauthoryear{Bonab and Can}{Bonab and Can}{2016}]%
        {bib:bonabcikm}
{Hamed~R. Bonab} {and} {Fazli Can}. 2016.
\newblock \showarticletitle{A theoretical framework on the ideal number of
  classifiers for online ensembles in data streams}. In {\em Proceedings of the
  25th {ACM} International Conference on Information and Knowledge Management,
  {CIKM} 2016, Indianapolis, IN, USA, October 24-28, 2016}. 2053--2056.
\newblock


\bibitem[\protect\citeauthoryear{Brzezinski and Stefanowski}{Brzezinski and
  Stefanowski}{2014a}]%
        {bib:oaue}
{Dariusz Brzezinski} {and} {Jerzy Stefanowski}. 2014a.
\newblock \showarticletitle{Combining block-based and online methods in
  learning ensembles from concept drifting data streams}.
\newblock {\em Information Sciences\/}  {265} (2014), 50--67.
\newblock


\bibitem[\protect\citeauthoryear{Brzezinski and Stefanowski}{Brzezinski and
  Stefanowski}{2014b}]%
        {bib:aue}
{Dariusz Brzezinski} {and} {Jerzy Stefanowski}. 2014b.
\newblock \showarticletitle{Reacting to different types of concept drift: The
  accuracy updated ensemble algorithm}.
\newblock {\em {IEEE} Transactions on Neural Networks and Learning Systems\/}
  {25}, 1 (2014), 81--94.
\newblock


\bibitem[\protect\citeauthoryear{Chan}{Chan}{1999}]%
        {bib:chan1999weighted}
{Lai-Wan Chan}. 1999.
\newblock \showarticletitle{Weighted least square ensemble networks}. In {\em
  International Joint Conference on Neural Networks (IJCNN 1999)}, Vol.~2.
  IEEE, 1393--1396.
\newblock


\bibitem[\protect\citeauthoryear{Conover}{Conover}{1999}]%
        {bib:statBook}
{W. Conover}. 1999.
\newblock {\em Practical~Nonparametric~Statistics}.
\newblock John Wiley~\&~Sons, New York.
\newblock


\bibitem[\protect\citeauthoryear{Dem{\v{s}}ar}{Dem{\v{s}}ar}{2006}]%
        {bib:statJournal}
{Janez Dem{\v{s}}ar}. 2006.
\newblock \showarticletitle{Statistical comparisons of classifiers over
  multiple data sets}.
\newblock {\em Journal of Machine Learning Research\/}  {7} (2006), 1--30.
\newblock


\bibitem[\protect\citeauthoryear{Dietterich and Bakiri}{Dietterich and
  Bakiri}{1995}]%
        {bib:output}
{Thomas~G. Dietterich} {and} {Ghulum Bakiri}. 1995.
\newblock \showarticletitle{Solving multiclass learning problems via
  error-correcting output codes}.
\newblock {\em Journal of Artificial Intelligence Research {(JAIR)}\/}  {2}
  (1995), 263--286.
\newblock


\bibitem[\protect\citeauthoryear{Ditzler and Polikar}{Ditzler and
  Polikar}{2013}]%
        {bib:nseimb}
{Gregory Ditzler} {and} {Robi Polikar}. 2013.
\newblock \showarticletitle{Incremental learning of concept drift from
  streaming imbalanced data}.
\newblock {\em {IEEE} Transactions on Knowledge and Data Engineering\/} {25},
  10 (2013), 2283--2301.
\newblock


\bibitem[\protect\citeauthoryear{Domingos and Hulten}{Domingos and
  Hulten}{2000}]%
        {bib:ht}
{Pedro~M. Domingos} {and} {Geoff Hulten}. 2000.
\newblock \showarticletitle{Mining high-speed data streams}. In {\em Sixth
  {ACM} {SIGKDD} International Conference on Knowledge Discovery and Data
  Mining, Boston, MA, USA, August 20-23, 2000}. 71--80.
\newblock


\bibitem[\protect\citeauthoryear{Elwell and Polikar}{Elwell and
  Polikar}{2011}]%
        {bib:nse}
{Ryan Elwell} {and} {Robi Polikar}. 2011.
\newblock \showarticletitle{Incremental learning of concept drift in
  nonstationary environments}.
\newblock {\em {IEEE} Transactions on Neural Networks and Learning Systems\/}
  {22}, 10 (2011), 1517--1531.
\newblock


\bibitem[\protect\citeauthoryear{Farid, Zhang, Hossain, Rahman, Strachan,
  Sexton, and Dahal}{Farid et~al\mbox{.}}{2013}]%
        {farid2013adaptive}
{Dewan~Md Farid}, {Li Zhang}, {Alamgir Hossain}, {Chowdhury~Mofizur Rahman},
  {Rebecca Strachan}, {Graham Sexton}, {and} {Keshav Dahal}. 2013.
\newblock \showarticletitle{An adaptive ensemble classifier for mining concept
  drifting data streams}.
\newblock {\em Expert Systems with Applications\/} {40}, 15 (2013), 5895--5906.
\newblock


\bibitem[\protect\citeauthoryear{Freund and Schapire}{Freund and
  Schapire}{1997}]%
        {bib:hedge}
{Yoav Freund} {and} {Robert~E. Schapire}. 1997.
\newblock \showarticletitle{A decision-theoretic generalization of on-line
  learning and an application to boosting}.
\newblock {\it J. Comput. System Sci.} {55}, 1 (1997), 119--139.
\newblock


\bibitem[\protect\citeauthoryear{Friedman}{Friedman}{2002}]%
        {bib:friedman}
{Jerome~H Friedman}. 2002.
\newblock \showarticletitle{Stochastic gradient boosting}.
\newblock {\em Computational Statistics \& Data Analysis\/} {38}, 4 (2002),
  367--378.
\newblock


\bibitem[\protect\citeauthoryear{Gama}{Gama}{2010}]%
        {bib:gama2010book}
{Joao Gama}. 2010.
\newblock {\em Knowledge discovery from data streams}.
\newblock CRC Press.
\newblock


\bibitem[\protect\citeauthoryear{Gama, Sebasti{\~a}o, and Rodrigues}{Gama
  et~al\mbox{.}}{2013}]%
        {bib:gama2013evaluating}
{Jo{\~a}o Gama}, {Raquel Sebasti{\~a}o}, {and} {Pedro~Pereira Rodrigues}. 2013.
\newblock \showarticletitle{On evaluating stream learning algorithms}.
\newblock {\em Machine Learning\/} {90}, 3 (2013), 317--346.
\newblock


\bibitem[\protect\citeauthoryear{Gama, Zliobaite, Bifet, Pechenizkiy, and
  Bouchachia}{Gama et~al\mbox{.}}{2014}]%
        {bib:survey}
{Jo{\~{a}}o Gama}, {Indre Zliobaite}, {Albert Bifet}, {Mykola Pechenizkiy},
  {and} {Abdelhamid Bouchachia}. 2014.
\newblock \showarticletitle{A survey on concept drift adaptation}.
\newblock {\it Comput. Surveys} {46}, 4 (2014), 44:1--44:37.
\newblock


\bibitem[\protect\citeauthoryear{Gao, Fan, and Han}{Gao et~al\mbox{.}}{2007}]%
        {bib:gao}
{Jing Gao}, {Wei Fan}, {and} {Jiawei Han}. 2007.
\newblock \showarticletitle{On appropriate assumptions to mine data streams:
  Analysis and practice}. In {\em Seventh IEEE International Conference on Data
  Mining (ICDM 2007)}. IEEE, 143--152.
\newblock


\bibitem[\protect\citeauthoryear{Gomes, Barddal, Enembreck, and Bifet}{Gomes
  et~al\mbox{.}}{2017}]%
        {bib:surveybifet2017}
{Heitor~Murilo Gomes}, {Jean~Paul Barddal}, {Fabr\'{\i}cio Enembreck}, {and}
  {Albert Bifet}. 2017.
\newblock \showarticletitle{A survey on ensemble learning for data stream
  classification}.
\newblock {\it Comput. Surveys} {50}, 2, Article 23 (March 2017), 36 pages.
\newblock
\showISSN{0360-0300}


\bibitem[\protect\citeauthoryear{Han, Zhang, and Wang}{Han
  et~al\mbox{.}}{2015}]%
        {bib:han2015}
{Dong-Hong Han}, {Xin Zhang}, {and} {Guo-Ren Wang}. 2015.
\newblock \showarticletitle{Classifying uncertain and evolving data streams
  with distributed extreme learning machine}.
\newblock {\em Journal of Computer Science and Technology\/} {30}, 4 (2015),
  874--887.
\newblock


\bibitem[\protect\citeauthoryear{Hansen, Pereyra, and Scherer}{Hansen
  et~al\mbox{.}}{2013}]%
        {bib:lsqbook}
{Per~Christian Hansen}, {V{\'\i}ctor Pereyra}, {and} {Godela Scherer}. 2013.
\newblock {\em Least squares data fitting with applications}.
\newblock Johns Hopkins University Press, Baltimore, Md.
\newblock


\bibitem[\protect\citeauthoryear{Hern{\'a}ndez-Lobato, Mart{\'\i}Nez-Mu{\~n}Oz,
  and Su{\'a}rez}{Hern{\'a}ndez-Lobato et~al\mbox{.}}{2013}]%
        {bib:hernandez2013large}
{Daniel Hern{\'a}ndez-Lobato}, {Gonzalo Mart{\'\i}Nez-Mu{\~n}Oz}, {and}
  {Alberto Su{\'a}rez}. 2013.
\newblock \showarticletitle{How large should ensembles of classifiers be?}
\newblock {\em Pattern Recognition\/} {46}, 5 (2013), 1323--1336.
\newblock


\bibitem[\protect\citeauthoryear{Hulten, Spencer, and Domingos}{Hulten
  et~al\mbox{.}}{2001}]%
        {bib:rothyper}
{Geoff Hulten}, {Laurie Spencer}, {and} {Pedro~M. Domingos}. 2001.
\newblock \showarticletitle{Mining time-changing data streams}. In {\em Seventh
  {ACM} {SIGKDD} International Conference on Knowledge Discovery and Data
  Mining, San Francisco, CA, USA, August 26-29, 2001}. 97--106.
\newblock


\bibitem[\protect\citeauthoryear{Kolter and Maloof}{Kolter and Maloof}{2003}]%
        {bib:dwm2}
{Jeremy~Z Kolter} {and} {Marcus~A Maloof}. 2003.
\newblock \showarticletitle{Dynamic weighted majority: A new ensemble method
  for tracking concept drift}. In {\em Third IEEE International Conference on
  Data Mining (ICDM 2003)}. IEEE, 123--130.
\newblock


\bibitem[\protect\citeauthoryear{Kolter and Maloof}{Kolter and Maloof}{2005}]%
        {bib:addExpert}
{Jeremy~Z. Kolter} {and} {Marcus~A. Maloof}. 2005.
\newblock \showarticletitle{Using additive expert ensembles to cope with
  concept drift}. In {\em Proceedings of 22nd International Conference on
  Machine Learning, {(ICML 2005)}, Bonn, Germany, August 7-11, 2005}. 449--456.
\newblock


\bibitem[\protect\citeauthoryear{Kolter and Maloof}{Kolter and Maloof}{2007}]%
        {bib:dwm}
{J~Zico Kolter} {and} {Marcus~A Maloof}. 2007.
\newblock \showarticletitle{Dynamic weighted majority: An ensemble method for
  drifting concepts}.
\newblock {\em Journal of Machine Learning Research\/}  {8} (2007), 2755--2790.
\newblock


\bibitem[\protect\citeauthoryear{Krawczyk, Minku, Gama, Stefanowski, and
  Wo{\'z}niak}{Krawczyk et~al\mbox{.}}{2017}]%
        {bib:surveygama2017}
{Bartosz Krawczyk}, {Leandro~L Minku}, {Jo{\~a}o Gama}, {Jerzy Stefanowski},
  {and} {Micha{\l} Wo{\'z}niak}. 2017.
\newblock \showarticletitle{Ensemble learning for data stream analysis: A
  survey}.
\newblock {\em Information Fusion\/}  {37} (2017), 132--156.
\newblock


\bibitem[\protect\citeauthoryear{Kuncheva}{Kuncheva}{2004}]%
        {bib:kuncheva1}
{Ludmila~I. Kuncheva}. 2004.
\newblock \showarticletitle{Classifier ensembles for changing environments}. In
  {\em Proceedings of the 5th International Workshop on Multiple Classifier
  Systems, {(MCS 2004)}, Cagliari, Italy, June 9-11, 2004}. 1--15.
\newblock


\bibitem[\protect\citeauthoryear{Kuncheva}{Kuncheva}{2008}]%
        {bib:kuncheva2}
{Ludmila~I Kuncheva}. 2008.
\newblock \showarticletitle{Classifier ensembles for detecting concept change
  in streaming data: Overview and perspectives}. In {\em 2nd Workshop SUEMA},
  Vol. 2008. 5--10.
\newblock


\bibitem[\protect\citeauthoryear{Latinne, Debeir, and Decaestecker}{Latinne
  et~al\mbox{.}}{2001}]%
        {bib:latinne2001limiting}
{Patrice Latinne}, {Olivier Debeir}, {and} {Christine Decaestecker}. 2001.
\newblock \showarticletitle{Limiting the number of trees in random forests}. In
  {\em International Workshop on Multiple Classifier Systems}. Springer,
  178--187.
\newblock


\bibitem[\protect\citeauthoryear{Littlestone}{Littlestone}{1987}]%
        {bib:winnow}
{Nick Littlestone}. 1987.
\newblock \showarticletitle{Learning quickly when irrelevant attributes abound:
  {A} new linear-threshold algorithm}.
\newblock {\em Machine Learning\/} {2}, 4 (1987), 285--318.
\newblock


\bibitem[\protect\citeauthoryear{Littlestone and Warmuth}{Littlestone and
  Warmuth}{1994}]%
        {bib:wm}
{Nick Littlestone} {and} {Manfred~K. Warmuth}. 1994.
\newblock \showarticletitle{The weighted majority algorithm}.
\newblock {\em Information and Computation\/} {108}, 2 (1994), 212--261.
\newblock


\bibitem[\protect\citeauthoryear{Masud, Gao, Khan, Han, and
  Thuraisingham}{Masud et~al\mbox{.}}{2011}]%
        {bib:novelclass}
{Mohammad~M. Masud}, {Jing Gao}, {Latifur Khan}, {Jiawei Han}, {and}
  {Bhavani~M. Thuraisingham}. 2011.
\newblock \showarticletitle{Classification and novel class detection in
  concept-drifting data streams under time constraints}.
\newblock {\em {IEEE} Transactions on Knowledge and Data Engineering\/} {23}, 6
  (2011), 859--874.
\newblock


\bibitem[\protect\citeauthoryear{Minku, White, and Yao}{Minku
  et~al\mbox{.}}{2010}]%
        {bib:minku1}
{Leandro~L. Minku}, {Allan~P. White}, {and} {Xin Yao}. 2010.
\newblock \showarticletitle{The impact of diversity on online ensemble eearning
  in the presence of concept drift}.
\newblock {\em {IEEE} Transactions on Knowledge and Data Engineering\/} {22}, 5
  (2010), 730--742.
\newblock


\bibitem[\protect\citeauthoryear{Minku and Yao}{Minku and Yao}{2012}]%
        {bib:minku2}
{Leandro~L. Minku} {and} {Xin Yao}. 2012.
\newblock \showarticletitle{{DDD:} {A} new ensemble approach for dealing with
  concept drift}.
\newblock {\em {IEEE} Transactions on Knowledge and Data Engineering\/} {24}, 4
  (2012), 619--633.
\newblock


\bibitem[\protect\citeauthoryear{Mustafa, Haque, Khan, Baron, and
  Thuraisingham}{Mustafa et~al\mbox{.}}{2014}]%
        {bib:mustafa}
{Ahmad Mustafa}, {Ahsanul Haque}, {Latifur Khan}, {Michael Baron}, {and}
  {Bhavani Thuraisingham}. 2014.
\newblock \showarticletitle{Evolving stream classification using change
  detection}. In {\em International Conference on Collaborative Computing:
  Networking, Applications and Worksharing (CollaborateCom 2014)}. IEEE,
  154--162.
\newblock


\bibitem[\protect\citeauthoryear{Nishida, Yamauchi, and Omori}{Nishida
  et~al\mbox{.}}{2005}]%
        {bib:ace}
{Kyosuke Nishida}, {Koichiro Yamauchi}, {and} {Takashi Omori}. 2005.
\newblock \showarticletitle{ACE: Adaptive classifiers-ensemble system for
  concept-drifting environments}.
\newblock In {\em Multiple Classifier Systems}. Springer, 176--185.
\newblock


\bibitem[\protect\citeauthoryear{Oshiro, Perez, and Baranauskas}{Oshiro
  et~al\mbox{.}}{2012}]%
        {bib:oshiro2012many}
{Thais~Mayumi Oshiro}, {Pedro~Santoro Perez}, {and} {Jos{\'e}~Augusto
  Baranauskas}. 2012.
\newblock \showarticletitle{How many trees in a random forest?}. In {\em
  International Workshop on Machine Learning and Data Mining in Pattern
  Recognition}. Springer, 154--168.
\newblock


\bibitem[\protect\citeauthoryear{Oza}{Oza}{2001}]%
        {bib:ozaphd}
{Nikunj~C. Oza}. 2001.
\newblock {\em Online Ensemble Learning}.
\newblock Ph.D. Dissertation. Computer Science Division, Univ. California,
  Berkeley, CA, USA.
\newblock


\bibitem[\protect\citeauthoryear{Oza and Russell}{Oza and Russell}{2001}]%
        {bib:oza2}
{Nikunj~C Oza} {and} {Stuart Russell}. 2001.
\newblock \showarticletitle{Experimental comparisons of online and batch
  versions of bagging and boosting}. In {\em Proceedings of the Seventh {ACM}
  {SIGKDD} International Conference on Knowledge Discovery and Data Mining, San
  Francisco, CA, USA, August 26-29, 2001}. ACM, 359--364.
\newblock


\bibitem[\protect\citeauthoryear{Saari}{Saari}{2008}]%
        {bib:saari2008}
{Donald~G. Saari}. 2008.
\newblock \showarticletitle{Complexity and the geometry of voting}.
\newblock {\em Mathematical and Computer Modelling\/} {48}, 9-10 (2008),
  1335--1356.
\newblock


\bibitem[\protect\citeauthoryear{Scheffe}{Scheffe}{1959}]%
        {bib:scheffe}
{Henry Scheffe}. 1959.
\newblock {\em The Analysis of Variance}.
\newblock John Wiley, New York.
\newblock


\bibitem[\protect\citeauthoryear{Street and Kim}{Street and Kim}{2001}]%
        {bib:sea}
{W.~Nick Street} {and} {YongSeog Kim}. 2001.
\newblock \showarticletitle{A streaming ensemble algorithm {(SEA)} for
  large-scale classification}. In {\em Proceedings of the Seventh {ACM}
  {SIGKDD} International Conference on Knowledge Discovery and Data Mining, San
  Francisco, CA, USA, August 26-29, 2001}. 377--382.
\newblock


\bibitem[\protect\citeauthoryear{Sun, Tang, Minku, Wang, and Yao}{Sun
  et~al\mbox{.}}{2016}]%
        {bib:sun2016}
{Yu Sun}, {Ke Tang}, {Leandro~L Minku}, {Shuo Wang}, {and} {Xin Yao}. 2016.
\newblock \showarticletitle{Online ensemble learning of data streams with
  gradually evolved classes}.
\newblock {\em IEEE Transactions on Knowledge and Data Engineering\/} {28}, 6
  (2016), 1532--1545.
\newblock


\bibitem[\protect\citeauthoryear{Tumer and Ghosh}{Tumer and Ghosh}{1996}]%
        {bib:decbound}
{Kagan Tumer} {and} {Joydeep Ghosh}. 1996.
\newblock \showarticletitle{Analysis of decision boundaries in linearly
  combined neural classifiers}.
\newblock {\em Pattern Recognition\/} {29}, 2 (1996), 341--348.
\newblock


\bibitem[\protect\citeauthoryear{Wang, Fan, Yu, and Han}{Wang
  et~al\mbox{.}}{2003}]%
        {bib:awe}
{Haixun Wang}, {Wei Fan}, {Philip~S. Yu}, {and} {Jiawei Han}. 2003.
\newblock \showarticletitle{Mining concept-drifting data streams using ensemble
  classifiers}. In {\em Proceedings of the Ninth {ACM} {SIGKDD} International
  Conference on Knowledge Discovery and Data Mining, Washington, DC, USA,
  August 24 - 27, 2003}. 226--235.
\newblock


\bibitem[\protect\citeauthoryear{Wang, Minku, and Yao}{Wang
  et~al\mbox{.}}{2015}]%
        {bib:wang2015}
{Shuo Wang}, {Leandro~L Minku}, {and} {Xin Yao}. 2015.
\newblock \showarticletitle{Resampling-based ensemble methods for online class
  imbalance learning}.
\newblock {\em IEEE Transactions on Knowledge and Data Engineering\/} {27}, 5
  (2015), 1356--1368.
\newblock


\bibitem[\protect\citeauthoryear{Zamani, Beigy, and Shaban}{Zamani
  et~al\mbox{.}}{2016}]%
        {bib:zamani}
{Mohammadzaman Zamani}, {Hamid Beigy}, {and} {Amirreza Shaban}. 2016.
\newblock \showarticletitle{Cascading randomized weighted majority: {a} new
  online ensemble learning algorithm}.
\newblock {\em Intelligent Data Analysis\/} {20}, 4 (2016), 877--889.
\newblock


\bibitem[\protect\citeauthoryear{Zhang, Zhu, and Shi}{Zhang
  et~al\mbox{.}}{2008}]%
        {bib:zhang}
{Peng Zhang}, {Xingquan Zhu}, {and} {Yong Shi}. 2008.
\newblock \showarticletitle{Categorizing and mining concept drifting data
  streams}. In {\em Proceedings of the 14th {ACM} {SIGKDD} International
  Conference on Knowledge Discovery and Data Mining, Las Vegas, Nevada, USA,
  August 24-27, 2008}. 812--820.
\newblock


\bibitem[\protect\citeauthoryear{Zhu, Zhang, Lin, and Shi}{Zhu
  et~al\mbox{.}}{2010}]%
        {bib:sampling}
{Xingquan Zhu}, {Peng Zhang}, {Xiaodong Lin}, {and} {Yong Shi}. 2010.
\newblock \showarticletitle{Active learning from stream data using optimal
  weight classifier ensemble}.
\newblock {\em {IEEE} Transactions on Systems, Man, and Cybernetics, Part
  {B}\/} {40}, 6 (2010), 1607--1621.
\newblock


\bibitem[\protect\citeauthoryear{Zliobaite}{Zliobaite}{2013}]%
        {bib:electcritic}
{Indre Zliobaite}. 2013.
\newblock \showarticletitle{How good is the electricity benchmark for
  evaluating concept drift adaptation}.
\newblock {\em CoRR\/}  {abs/1301.3524} (2013).
\newblock


\end{thebibliography}

\received{July 2016}{May 2017}{August 2017}

\end{document}